\documentclass{article}

\usepackage[preprint]{corl_2025} 

\usepackage{etoolbox} 
\usepackage[utf8]{inputenc} 
\usepackage[T1]{fontenc}    
\usepackage{fonttable}
\usepackage[english]{babel} 
\usepackage[protrusion=true,expansion=true]{microtype}
\usepackage{comment}
\usepackage{cancel}
\usepackage{csquotes}
\usepackage{listings}
\usepackage{nameref}
\usepackage{paralist}
\usepackage{xspace}
\usepackage{setspace}
\usepackage{makeidx}
\usepackage{pdfpages}
\usepackage{times}

\usepackage{amsmath,amssymb} 
\usepackage{amsfonts}       
\usepackage{mathtools}
\usepackage{array} 
\usepackage{nicefrac}       
\usepackage{breqn}          
\usepackage{textcomp}
\usepackage{bm} 
\usepackage{pifont}
\usepackage[
  separate-uncertainty = true,
  multi-part-units = repeat
]{siunitx}
\usepackage{mathrsfs}

\usepackage{graphicx}
\usepackage{wrapfig}
\usepackage{adjustbox} 
\usepackage{epsfig}

\usepackage{float}
\usepackage{subcaption}
\usepackage[font=small,labelfont=bf]{caption}
\usepackage{lscape}      

\usepackage{tikz}  
\usepackage{makecell}
\usepackage{overpic}

\newsavebox{\subfigbox}

\usepackage{algorithm}
\usepackage[noend]{algpseudocode}

\usepackage{array} 
\usepackage{tabularx}
\usepackage{multirow}
\usepackage{multicol}
\usepackage{booktabs}   
\usepackage{tablefootnote}

\usepackage{paralist}
\usepackage{enumitem}
\setlist[itemize]{noitemsep,leftmargin=*,topsep=0in}
\setlist[enumerate]{noitemsep,leftmargin=*,topsep=0in}

\usepackage{url}            
\PassOptionsToPackage{table}{xcolor}
\usepackage{color,colortbl} 
\usepackage{soul}

\PassOptionsToPackage{capitalize, noabbrev}{cleveref}

\hypersetup{pagebackref=false,breaklinks=true,colorlinks,urlcolor=blue,citecolor=blue,linkcolor=blue,bookmarks=false}

\newcommand{\Eq}[1]{Eq.~\ref{eq:#1}}

\usepackage{blindtext}
\usepackage{bbm}

\usepackage{lipsum}

\usepackage{titlesec}
\titlespacing{\section}{0pt}{0.3\baselineskip}{0.25\baselineskip}
\titlespacing{\subsection}{0pt}{0.2\baselineskip}{0.15\baselineskip}
\titlespacing{\subsubsection}{0pt}{0.05\baselineskip}{0.03\baselineskip}

\renewcommand{\paragraph}[1]{\vspace{0.2em}\noindent\textit{#1} --}

\definecolor{color1}{rgb}{.6,.4,.05}
\definecolor{color2}{rgb}{0,.7,.7}
\definecolor{color3}{rgb}{0.35,0.75,0.0}
\definecolor{color4}{rgb}{0.4,0.8,0}
\definecolor{color5}{rgb}{0.5,0.0,0.5}
\definecolor{revision_color}{rgb}{1,0,0}

\newcommand{\cmark}{\ding{51}} 
\newcommand{\xmark}{\ding{55}} 

\newcommand{\modelname}{\textsc{Amplify}\xspace}

\title{\modelname: Actionless Motion Priors for\\Robot Learning from Videos}

\author{%
  Jeremy A. Collins$^{1}$\thanks{Equal contribution.\\ $^{1}$Georgia Tech $^{2}$Georgia Tech Research Institute}, %
  Loránd Cheng$^{1}$\footnotemark[1], %
  Kunal Aneja$^{1}$, %
  Albert Wilcox$^{1}$,\\%
 \textbf{ Benjamin Joffe$^{1,2}$}, %
  \textbf{Animesh Garg$^{1}$}%
}

\date{} 

\begin{document}
\maketitle




\vspace{-5mm}
\begin{center}
    \centering
    \captionsetup{type=figure}
    \includegraphics[width=\textwidth]{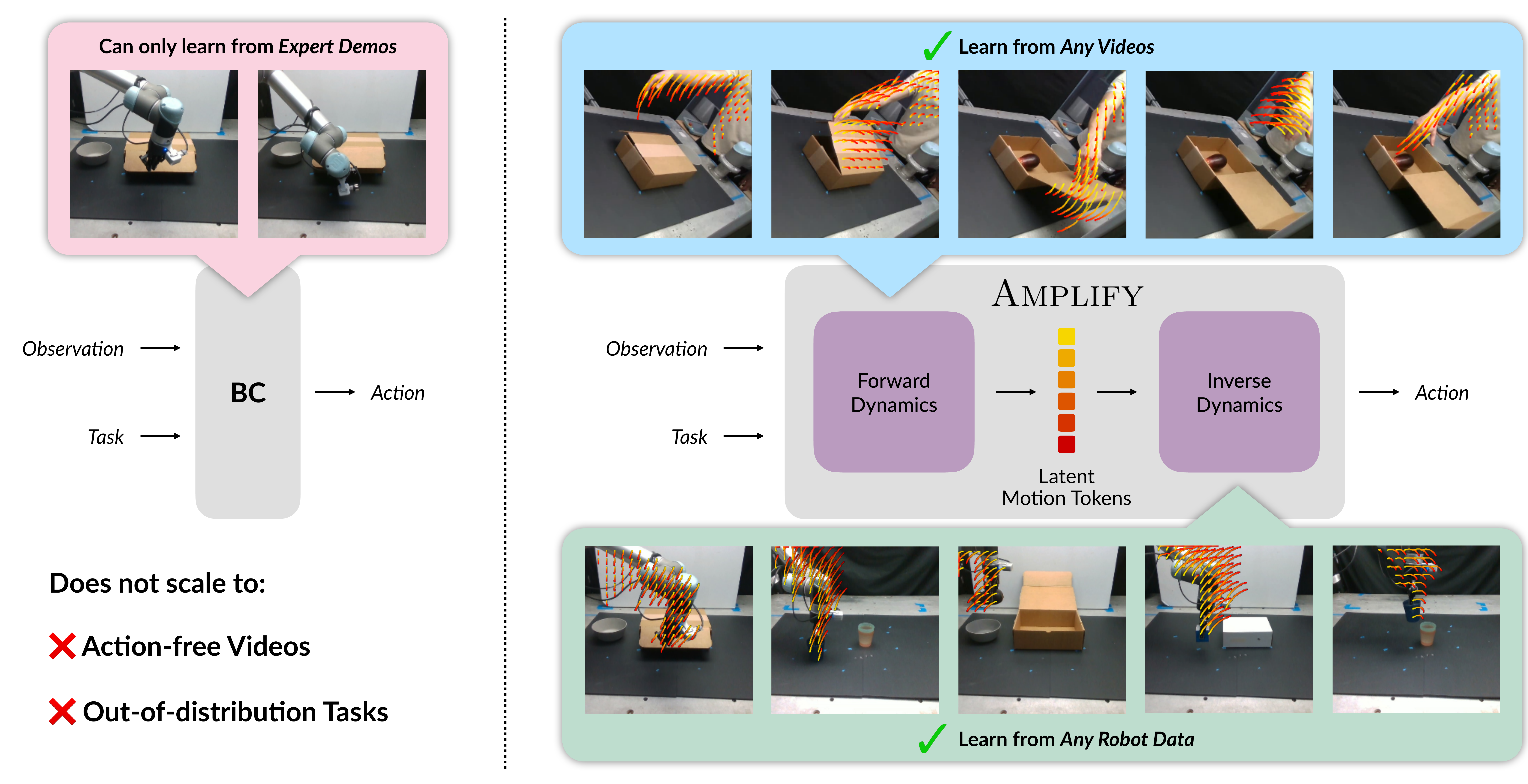}
    \vspace{-5mm}
    \caption{Overview. \modelname decomposes policy learning into forward and inverse dynamics, using latent keypoint motion as an intermediate representation. The forward model can be trained on \textit{any} video data, while the inverse model can be trained \textit{any} interaction data. In contrast with behavior cloning (BC), \modelname requires fewer demos, can generalize to tasks for which we have \textit{zero} action data, and learn from human videos.
    }
    \label{fig:banner}
\end{center}
\vspace{-3mm}


\begin{abstract}

Action-labeled data for robotics is scarce and expensive, limiting the generalization of learned policies. In contrast, vast amounts of action-free video data are readily available, but translating these observations into effective policies remains a challenge. We introduce \modelname, a novel framework that leverages large-scale video data by encoding visual dynamics into compact, discrete motion tokens derived from keypoint trajectories. Our modular approach separates visual motion prediction from action inference, decoupling the challenges of learning \textbf{\textit{what}} motion defines a task from \textbf{\textit{how}} robots can perform it. We train a forward dynamics model on abundant action-free videos and an inverse dynamics model on a limited set of action-labeled examples, allowing for independent scaling. Extensive evaluations demonstrate that the learned dynamics are both accurate—achieving up to 3.7× better MSE and over 2.5× better pixel prediction accuracy compared to prior approaches—and broadly useful. In downstream policy learning, our dynamics predictions enable a 1.2-2.2× improvement in low-data regimes, a 1.4× average improvement by learning from action-free human videos, and the first generalization to LIBERO tasks from zero in-distribution action data. Beyond robotic control, we find the dynamics learned by \modelname to be a versatile latent world model, enhancing video prediction quality. Our results present a novel paradigm leveraging heterogeneous data sources to build efficient, generalizable world models. More information can be found at \href{https://amplify-robotics.github.io/}{amplify-robotics.github.io}.

\end{abstract}



\section{Introduction}
\label{sec:intro}
\vspace{-2mm}
Recent successes in harnessing internet-scale data to train image and language foundation models~\citep{radford2019language, brown2020language, achiam2023gpt, radford2021learningtransferablevisualmodels, ramesh2022hierarchicaltextconditionalimagegeneration, rombach2022highresolutionimagesynthesislatent} have spurred an analogous push in robotics. 
In contrast with earlier methods that focused on achieving expert-level capabilities in narrow, controlled domains, recent efforts in robotics have aimed to generalize across tasks, object categories, object instances, environments, and the abundant variety of conditions present in the natural world~\citep{brohan_rt-1_nodate,brohan2023rt,padalkar2023open,black2024pi0,pertsch2025fast,shi2025hi}. 
However, in order to train such generalist models, the typical behavior cloning (BC) approach requires prohibitively large amounts of \textbf{action-labeled expert demonstrations}. Datasets that are considered large-scale for robotics~\citep{brohan_rt-1_nodate,padalkar2023open,khazatsky2024droid} take weeks or months to collect a few \textit{hundred} hours of interaction data, falling far short of the roughly \textit{one billion} hours of video data available on the internet. 
Therefore, methods that incorporate large-scale pre-training on these more abundant modalities tend to generalize better from limited action data~\citep{kim_openvla_2024,black__0_2024,brohan2023rt}. Videos, in particular, contain rich priors on temporally-extended dynamics, behaviors, and semantics, which can be used to learn a predictive model of the world~\citep{du_learning_2023,mccarthy_towards_2024,rybkin_learning_2019,nvidia_cosmos_2025,ho2022imagenvideohighdefinition,kong2024hunyuanvideo,veo2,bar2024lumiere} 

Prior work has leveraged video pre-training to learn representations using a number of auxiliary tasks such as reward and value prediction~\citep{ma2022vip,ghosh_reinforcement_2023,bhateja_robotic_2023,dashora2025viva} or time-contrastive loss terms~\citep{ma2022vip,sermanet2018time,nair2022r3m}. While useful as representations, these methods only learn an encoder for static observations and do not explicitly model sequential dynamics. In contrast, model-based approaches can improve sample efficiency by separating the challenge of policy learning from learning dynamics \citep{moerland2023model}. Since videos contain rich priors over object and agent dynamics, model-based methods offer a promising avenue for learning from limited action data. One such approach is to train a full video prediction model to capture visual dynamics, which can act as a reference generator for downstream policies~\citep{du_learning_2023,hu2024video}. However, predicting in pixel space is computationally intensive and costly to run at high frequencies, forcing these methods to make compromises like open-loop control~\citep{du_learning_2023} or partial denoising~\citep{hu2024video}. As a result, a number of works have aimed to learn \textit{latent action} representations from videos using next-frame prediction~\citep{bruce2024genie,ye2024latent,chen_moto_2024} or latent consistency~\citep{cui2024dynamo}, efficiently modeling features that are predictive of the future. While this avoids high inference costs, these representations are still trained on image reconstruction/prediction objectives, capturing textural details or visually salient features that may not be relevant to policy learning.

Motivated by the desire to capture motion rather than appearance, optical flow and keypoint tracking have emerged as appealing abstractions for extracting action information from videos without action labels. Recent advances in computer vision have enabled efficient and precise pixel-level point tracking, even through occlusions and limited out-of-frame tracking \citep{karaev2023cotracker, wang2023omnimotion, doersch2023tapir, SpatialTracker}. As these capabilities enable fine-grained capture of motion and scene dynamics, they have found applications in robotics for visual imitation learning \citep{vecerik_robotap_2023} and tool use \citep{qin2020keto}. A number of prior works predict motion from images as optical flow \citep{yuan2024general,lin_flowretrieval_2024, Ko2023Learning} or by modeling the trajectories of specified keypoints \citep{kareer2024egomimic,gao_k-vil_2023,fang_keypoint_2024,gaoflip,manuelli2019kpam,guzey_bridging_2024,xu_flow_2024}. However, many of these works still rely on prohibitively expensive video prediction models \citep{xu_flow_2024,bharadhwaj_gen2act_2024,Ko2023Learning}, object-centric mask extraction \citep{xu_flow_2024,manuelli2019kpam,fang_keypoint_2024,bharadhwaj2024track2act}, calibrated cameras \citep{guzey_bridging_2024}, or inefficient online planning \citep{gaoflip}, limiting their generality. 

Two of the most general keypoint modeling approaches are ATM \citep{wen_any-point_2024} and Track2Act \citep{bharadhwaj2024track2act}, which aim to learn a universal keypoint dynamics model to predict the future trajectories of arbitrary points in an image, and condition a policy on these predictions. However, Track2Act relies on the often unrealistic assumption of a goal image and restricts its output space to single-object rigid-body transformations. ATM, while more flexible in its representation, relies on unrealistic point-sampling heuristics during training that cannot be replicated during inference. In addition, neither ATM nor Track2Act learn a latent space abstraction of keypoints, leaving them with high computational costs much like pixel-space video generation and potentially hindering generalization. Due to their high computational costs, Track2Act requires open-loop trajectory generation, and ATM only generates tracks for 32 points during policy inference, resulting in very coarse dynamics predictions. Further discussion and comparison to related work can be found in Appendix \ref{apdx:extended-related-work}.



In this paper, we investigate the use of \emph{latent} keypoint motion as an abstraction for learning valuable action priors from action-free video data, combining the benefits of latent dynamics prediction with the explicit motion information captured in keypoint trajectories. 
We propose \modelname: Actionless Motion Priors for Learning Inverse and Forward Dynamics, a three-stage framework that flexibly decouples dynamics modeling from policy learning. First, we learn a compact latent space for modeling the motion of a dense grid of keypoints. Second, we train a latent dynamics model to predict a sequence of latent motions based on the current observation. Finally, an inverse dynamics model learns to map predicted latent motions to low-level robot actions for execution. Notably, this modular approach allows the first two stages to be trained on \textit{any} video data, while the inverse dynamics policy can be trained on \textit{any} interaction data (Figure \ref{fig:banner}). We show that this has profound implications for policy generalization in Section \ref{sec:generalization}.

Through extensive real-world and simulated experiments, we evaluate both the accuracy and downstream utility of our latent dynamics model. Compared to state-of-the-art baselines, we observe that \modelname leads to improved keypoint trajectory prediction, lowering mean-squared error by over $3\times$. We then demonstrate that these predictions are useful for control; conditioning the inverse dynamics policy on latent motions is a valuable prior that allows for more data-efficient learning and generalization to tasks for which we have \textit{no action-labeled data}. Finally, we examine the versatility of our motion-based representations beyond control for tasks such as conditional video prediction.

In summary, we make the following \textit{key contributions}:
\begin{enumerate}
    \item We present the first \textit{latent} keypoint dynamics model and investigate crucial design choices.
    \item We demonstrate state-of-the-art keypoint prediction accuracy on three large-scale video datasets.
    \item We train a data-efficient and generalizable policy that can learn from action-free human data.
    \item We apply latent motions to conditional video generation, outperforming previous baselines.
\end{enumerate}


\section{\modelname: Method}
\label{method}
\vspace{-2mm}

\begin{figure*}[t]
    \centering
    \begin{subfigure}[b]{0.42\textwidth}
        \centering
        \includegraphics[width=\textwidth]{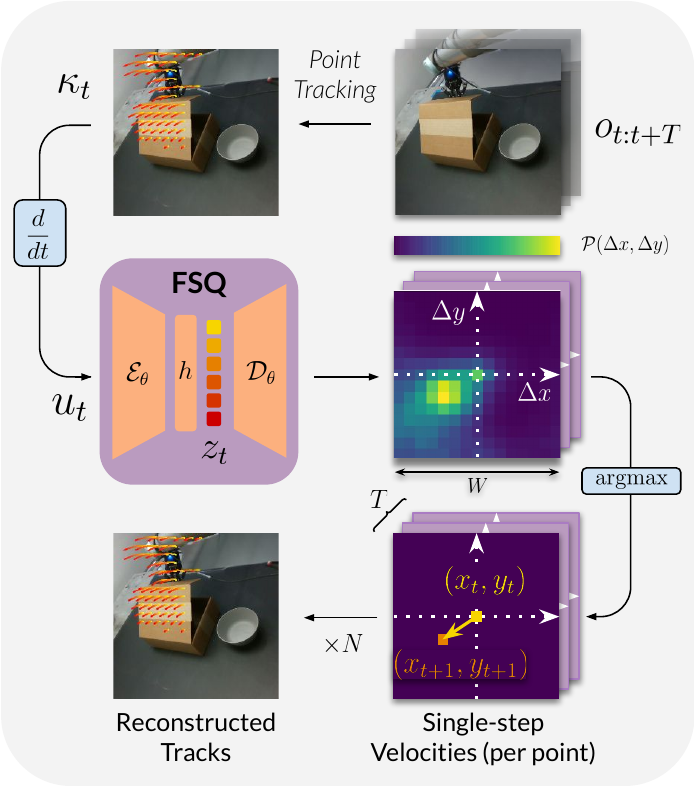}
        \caption{\small Motion Tokenization}
        \label{fig:tokenization}
    \end{subfigure}
    \begin{subfigure}[b]{0.564\textwidth}
        \begin{minipage}[b]{\textwidth}
            \centering
            \includegraphics[width=\textwidth]{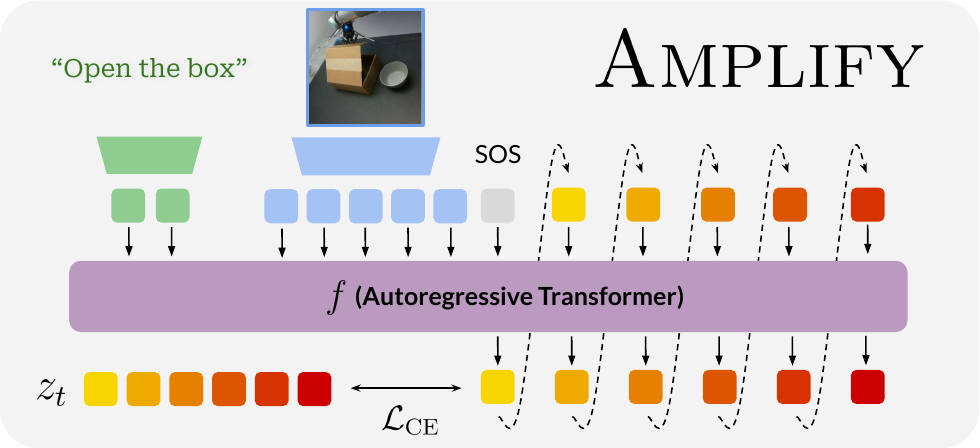}
            \caption{\small Forward Dynamics}
            \label{fig:forward}
        \end{minipage}
        \begin{minipage}[b]{\textwidth}
            \centering
            \includegraphics[width=\textwidth]{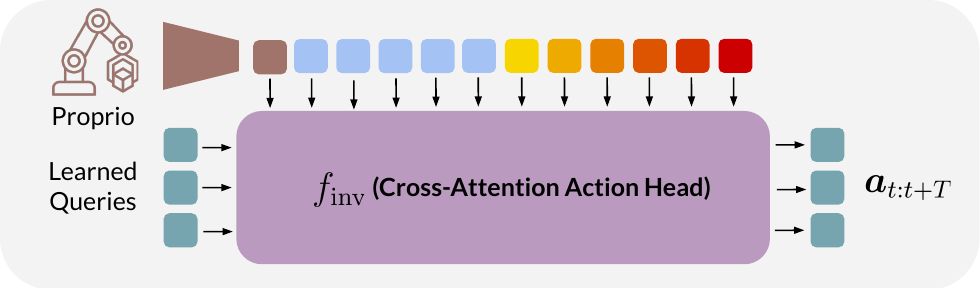}
            \caption{\small Inverse Dynamics}
            \label{fig:inverse}
        \end{minipage}
    \end{subfigure}
    \caption{Architecture. \modelname consists of a three-stage decomposition: (a) keypoint tracks are compressed into a discrete latent space using FSQ. For each timestep and each point, the decoder outputs a distribution in a local window centered around each point to reconstruct the instantaneous velocities, (b) a forward dynamics model is trained to predict the latent codes for the next $T$ timesteps given an input image and task description, and (c) an inverse dynamics model decodes predicted track tokens into an action chunk.}
    \label{fig:pipeline}
\end{figure*}

\paragraph{Problem Setup}
\label{sec:setup}
We assume access to two types of data: a video dataset $\mathcal{V} = \{(o_t,g)\}$ and a dataset of robot interaction data $\mathcal{R} = \{(o_t,q_t,a_t)\}$ where $o\in\mathcal{O}$ are RGB image observations, $g\in\mathcal{G}$ is a goal (e.g., a language description), and $a\in\mathcal{A},q\in\mathcal{Q}$ are the action and proprioceptive state of the robot, respectively\footnote{$\mathcal{V}$ and $\mathcal{R}$ need not be disjoint in general, and any goal-directed interaction data (demonstrations) may be included in both $\mathcal{V}$ and $\mathcal{R}$. However, $\mathcal{V}$ may additionally contain non-robot videos and $\mathcal{R}$ may contain undirected action data such as exploration or play data.}. Given these datasets, our aim is to learn the parameters of a visual control policy $\pi:\mathcal{O}\times\mathcal{Q}\times\mathcal{G} \to \mathcal{P}(\mathcal{A}) = f_{\text{inv}}(o_t,q_t,f(o_t, g))$ composed of a forward dynamics model $f: \mathcal{O}\times\mathcal{G}\to\mathcal{Z}$ that learns a \textit{motion prior} in a latent space $\mathcal{Z}$ and an inverse dynamics model $f_{\text{inv}}:\mathcal{O}\times\mathcal{Q}\times\mathcal{Z}\to\mathcal{A}$ that maps the latent motion to a sequence of actions. 
Crucially, this decomposition allows for independent scaling of $f$ and $f_{\text{inv}}$ by training on $\mathcal{V}$ and $\mathcal{R}$, respectively. We provide an extended discussion of the benefits of this decomposition in Appendix \ref{apdx:3-stage}. The following sections detail preprocessing (Sec. \ref{sec:preprocessing}), learning the latent motion representation (Sec. \ref{sec:vae}), and training the forward (Sec. \ref{sec:prediction}) and inverse (Sec. \ref{sec:inverse}) dynamics models.

\subsection{Preprocessing Keypoint Tracks}
\label{sec:preprocessing}
\vspace{-2mm}
We first augment $\mathcal{V}\to\mathcal{V}' = \{(o_t,\kappa_t,g)\}$ in a preprocessing step using the off-the-shelf point tracking model from \citep{karaev2023cotracker} to obtain a set of keypoint tracks $\kappa_t\in\mathbb{R}^{T\times N\times 2}$ for each timestep $t$. More precisely, we initialize a $20\times20$ uniform grid of $N=400$ points in each image $o_t$, then track the points through the next $T=16$ frames $o_{t:t+T}$, capturing their 2-dimensional pixel coordinates. Although extracting specific task-relevant keypoints could potentially yield more informative predictions, we favor the uniform grid for its simplicity and generality, similar to \citep{bharadhwaj2024track2act}, and find that it works effectively to model a variety of motions. Other works have attempted to select key points according to heuristics such as movement throughout the video \citep{wen_any-point_2024}, but we found that this led the model to learn spurious correlations and relies on unrealistic assumptions at test time. By reinitializing the grid of keypoints in each frame, we ensure no points are occluded and guarantee consistent coverage throughout every frame, even with moving cameras. See Appendix \ref{apdx:preprocessing} for further details on preprocessing.

\subsection{Motion Tokenization}
\label{sec:vae}
\vspace{-2mm}
Unlike prior keypoint-based methods which predict directly in pixel space \citep{wen_any-point_2024,bharadhwaj2024track2act,gaoflip,xu_flow_2024}, we argue that learning to predict dynamics in a compressed latent space enables a more efficient and generalizable representation, similar to findings in model-based reinforcement learning \citep{Hansen2022tdmpc,hansen_td-mpc2_2024,scannell2025discrete}. To this end, we learn a compact discrete latent space from pre-processed keypoint trajectories using Finite Scalar Quantization (FSQ) \citep{mentzer_finite_2023}, a drop-in replacement for vector-quantized variational autoencoders (VQ-VAEs) \citep{oord_neural_2018}. FSQ employs an implicit codebook and a single reconstruction loss term, avoiding representation collapse and resulting in better codebook utilization. 

Figure \ref{fig:tokenization} illustrates our tokenization scheme. We compute single-step velocities $u_t\in\mathbb{R}^{(T-1)\times N\times 2}$ from the pre-processed keypoint trajectories $\kappa_t$. Then, a keypoint encoder $\mathcal{E}_\theta:\mathbb{R}^{(T-1)\times N\times 2}\to\mathbb{R}^{b\times d}$ maps $u_t$ to a $d$-length sequence $\tilde{z}_t$ of latent vectors $\tilde{z}_{t,i} \in \mathbb{R}^b$, which are quantized via FSQ to a sequence $z_t \in \mathbb{Z}^{b\times d}$ of discrete codes, and decoded by the keypoint decoder $\mathcal{D}_\theta:\mathbb{R}^{b\times d}\to\mathbb{R}^{(T-1)\times N \times W^2}$ for reconstruction. Rather than just predicting the 2-dimensional pixel coordinate of each point directly, the decoder outputs a categorical distribution over $W^2$ classes representing a local $W\times W$ window of motions centered at the same point in the previous timestep. This imposes an inductive bias on the model toward next-keypoint predictions that are close to locations in the current timestep, and additionally better captures multimodal distributions compared to performing regression on the coordinates. The keypoint encoder has a causally-masked transformer encoder architecture, and the keypoint decoder is an unmasked transformer decoder that cross-attends between a sequence of $N$ learned positional encodings and the quantized codes from the encoder. The encoder and decoder are jointly trained on $\mathcal{V}$ using a cross-entropy loss:
\begin{equation} 
    \label{eq:autoencoder}
    \mathcal{L}_{AE}(\theta) = \text{CE}\Bigl(\mathcal{D}_\theta\Bigl(h\bigl(\mathcal{E}_\theta(u_t)\bigr)\Bigr),\, \omega_t \Bigr)
\end{equation}
where $\omega_t = \Omega(u_t)$, $\Omega:\mathbb{R}^{(T-1)\times N\times 2} \to \mathbb{R}^{(T-1)\times N\times W^2}$ maps ground-truth velocity vectors to their corresponding class based on the displacement in the local $W\times W$ window, and $h$ is the FSQ discretization function. When available, multi-view inputs are tokenized together into a single sequence of codes. For simplicity, we do not include the view dimension in our notation. For ablations and an extended discussion on the effects of these design choices, we refer readers to Appendix \ref{apdx:ablations}.

\subsection{Forward Dynamics (Actionless Motion Prior)}
\label{sec:prediction}
\vspace{-2mm}
After training the motion tokenizer, we train an autoregressive transformer $f(o_t,g)$ to predict the tokenized motion sequence $z_t$ corresponding to the video $o_{t:t+T}$ based on the current observation and task description. Image observations are encoded and projected into the embedding space of the transformer using the flattened feature map from a pre-trained ResNet-18 \citep{he_deep_2016} to generate $7\times7=49$ vision tokens per image. The summary token from a T5 \citep{raffel2020exploring} text embedding of the task description is used to tokenize language inputs. These conditioning tokens are then concatenated with a start of sequence (SOS) token and the latent motion tokens to predict the next tokens in the sequence (Figure \ref{fig:forward}). A block-causal attention mask is used, where the conditioning part of the sequence is non-causal and the motion tokens are causally masked. We use a cross-entropy loss on the predicted codes without decoding to full keypoint trajectories, and only back-propagate gradients to the dynamics model while the tokenizer remains frozen (Equation \ref{eq:dynamics}). \texttt{sg} refers to the stop-gradient operator.
\begin{equation}
    \label{eq:dynamics}
    \mathcal{L}_{\text{forward}} = \text{CE}\Bigl(f(o_t,g), \texttt{sg}(\mathcal{E}_\theta(u_t))\Bigr)
\end{equation}
\vspace{-5mm}

\subsection{Inverse Dynamics}
\label{sec:inverse}
\vspace{-2mm}
Finally, we learn an inverse dynamics model $f_{\text{inv}}(o_t,q_t,z_t)$ that decodes latent motion tokens into a distribution over action chunks $\bm{a}_t = a_{t:t+T}$, as shown in Figure~\ref{fig:inverse}. Importantly, this module is not conditioned on the goal and instead acts as a general reference follower trained on any interaction data $\mathcal{R}$. The model uses a transformer decoder with a sequence of learned tokens that cross-attend to image tokens, a linear projection of proprioceptive state, and codes from the motion tokenizer to produce a sequence of $d$ action tokens. These action tokens are fed into an action head to output a distribution over length-$T$ action chunks. Following BAKU \citep{haldar2024baku}, we opt for an isotropic Gaussian prior on the action distribution. In Appendix \ref{apdx:ablations-inverse}, we discuss alternative choices for the action head. The inverse dynamics model is trained with a negative log-likelihood (NLL) loss with a temporal discount $\gamma$ to reduce the impact of inaccurate predictions towards the end of the sequence.
\begin{equation}
    \label{eq:inv_dynamics}
    \mathcal{L}_{\text{inv}} = -\sum_{\tau=t}^{t+T-1} \gamma^{\tau - t} \cdot \log p\left(a_{\tau} \mid \mu_{\tau-t}, \sigma_{\tau-t}\right)
\end{equation}
where $\mu_{\tau-t} = f_{\text{inv}}^{\mu}(o_t,q_t,z_t)[\tau-t]$ and $\sigma_{\tau-t} = \exp(f_{\text{inv}}^{\sigma}(o_t,q_t,z_t)[\tau-t])$ are the predicted mean and standard deviation. The inverse dynamics model can be trained on ground truth tokens $z_t=\mathcal{E}_\theta(u_t)$, but in practice, we fine-tune the action decoder on the predicted outputs $\hat{z}_t$ of the forward dynamics model. Both the motion tokenizer and the forward dynamics model are frozen for this stage. The keypoint decoder $\mathcal{D}_\theta$ is not used, as we condition $f_{\text{inv}}$ on latent motions rather than decoded tracks.

\subsection{Inference}
\label{sec:inference}
\vspace{-2mm}
During inference, the forward dynamics model takes the current observation and task description at each timestep $t$ and autoregressively predicts a sequence of latent motion tokens $\hat{z}_t = f(o_t, g)$. The inverse dynamics model then decodes these tokens, along with image and proprioception tokens, into an action chunk $\bm{a}_t = f_{\text{inv}}(o_t, q_t, \hat{z}_t)$. Following ACT \citep{zhao_learning_nodate}, we use temporal ensembling to aggregate information over previously predicted action chunks using the same temporal discount $\gamma$.

\section{Experiments}
\label{experiments}
\vspace{-2mm}
We evaluate \modelname guided by two main axes of investigation: \textbf{quality} of dynamics prediction (Sec. \ref{exp:quality}) and \textbf{utility} of predictions for downstream tasks, including policy learning (Sec. \ref{exp:policy}) and conditional video generation (Sec. \ref{exp:video}). See Appendix \ref{apdx:experiment-training-details} for extended details on all experiments.


\begin{figure}[!t]
    \centering
    \includegraphics[width=\linewidth]{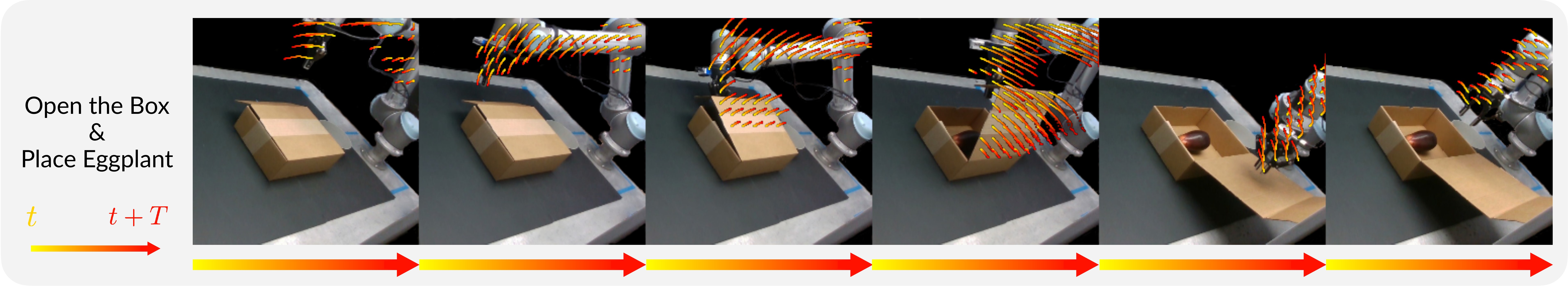}
    \caption{Decoded keypoint trajectory predictions from \modelname. Zero-movement points are not shown.}
    \label{fig:predictions}
\end{figure}

\subsection{Quality of Forward Dynamics Prediction}
\label{exp:quality}
\vspace{-2mm}
We test the prediction accuracy of our forward dynamics model on a combination of three simulated and real-world video datasets, including both human and robot data: BridgeData v2 \citep{walke2023bridgedata}, a large-scale robot dataset consisting of over 60k real-world rollouts of diverse manipulation tasks in 24 different environments; Something-Something v2 \citep{goyal2017somethingsomethingvideodatabase}, a video dataset consisting of over 220,000 videos of humans performing everyday manipulation tasks with a variety of objects and primitive motion categories; and LIBERO \citep{liu2023libero}, a benchmark of 130 diverse simulated robotic manipulation tasks, from which we use the observations from 6500 demonstration rollouts as a video dataset.

\begin{table}[!t]
    \centering
    \small
    \begin{minipage}{0.6\textwidth}
        \centering
        \begin{tabular}{l c c c}
            \toprule
            \textbf{Policy Learning} & \textbf{Forward} & \textbf{Inverse} & \textbf{BC} \\
            \textbf{Experiment} & \textbf{Dynamics} & \textbf{Dynamics} & \textbf{Baselines} \\
            \midrule
            In-Distribution    & $\mathcal{V}^R_{\text{id}}$ & $\mathcal{R}_{\text{id}}$ & $\mathcal{R}_{\text{id}}$ \\
            \rowcolor[HTML]{EFEFEF}
            Few-Shot           & $\mathcal{V}^R_{\text{id}}$ & $\subseteq\mathcal{R}_{\text{id}}$ & $\subseteq\mathcal{R}_{\text{id}}$ \\
            Cross-Embodiment   & $\mathcal{V}^R_{\text{id}} \cup \mathcal{V}^H_{\text{id}}$ & $\mathcal{R}_{\text{id}}$ & $\mathcal{R}_{\text{id}}$ \\
            \rowcolor[HTML]{EFEFEF}
            Generalization     & $\mathcal{V}^R_{\text{id}} \cup \mathcal{V}^R_{\text{ood}}$ & $\mathcal{R}_{\text{ood}}$ & $\mathcal{R}_{\text{ood}}$ \\
            \bottomrule
        \end{tabular}
    \end{minipage}%
    \hfill
    \begin{minipage}{0.39\textwidth}
        \caption{\textbf{Training dataset setup} for each component by experiment. Subscript \texttt{id} and \texttt{ood} indicate in-distribution and out of distribution tasks and superscript $H$ and $R$ distinguish human and robot video data. $\subseteq$ indicates training on limited subsets of the data.}
        \label{tab:experiment-notation}
    \end{minipage}
    \vspace{-5mm}
\end{table}

\begin{table}[!t]
\centering
\small
\caption{\textbf{Prediction}. \modelname achieves 3.7× better MSE and 2.5× better pixel accuracy compared to ATM, and a 4-6\% improvement over Track2Act, which uses a goal image, and Seer, which requires full video prediction.}
\label{tab:track-prediction}
\resizebox{0.99\linewidth}{!}{%
    \begin{tabular}{l c c c c c}
        \toprule
        \textbf{Method} & \multicolumn{3}{c}{LIBERO} & {BridgeDatav2} & {Something-Something v2}\\
        \cmidrule(lr){2-4} \cmidrule(lr){5-5} \cmidrule(lr){6-6} 
        Metric  & MSE $\downarrow$ & $\Delta_{\text{AUC}}$ $\uparrow$ & Pixel Acc. $\uparrow$ & $\Delta_{\text{AUC}}$ $\uparrow$ & $\Delta_{\text{AUC}}$ $\uparrow$  \\
        \midrule
        ATM \citep{wen_any-point_2024}                   & 0.022 & 0.767 & 0.250 & -- & -- \\
        \rowcolor[HTML]{EFEFEF}
        Track2Act \citep{bharadhwaj2024track2act}        & -- & -- & -- & 0.770 & 0.700 \\
        Seer \citep{gu2023seer}                          & -- & -- & -- & 0.914 & -- \\
        \rowcolor[HTML]{EFEFEF}
        \modelname                                       & \textbf{0.006} & \textbf{0.913} & \textbf{0.629} & \textbf{0.968} & \textbf{0.725} \\
        \bottomrule       
    \end{tabular}
}
\vspace{-5mm}
\end{table} 

We compare to ATM \citep{wen_any-point_2024} and Track2Act \citep{bharadhwaj2024track2act}, two state-of-the-art keypoint trajectory prediction approaches. In addition, on BridgeData v2 we compare track prediction accuracy to a baseline of first predicting videos with Seer \citep{gu2023seer}, then applying CoTracker \citep{karaev2023cotracker} to the initial set of points and tracking through the generated videos. Since our forward dynamics model predicts in latent space, we use the decoder from the Motion Tokenization stage for fair comparison in pixel space.  We measure performance on normalized tracks ($\kappa \in [-1,1]$) using Mean Squared Error (MSE), Pixel-Wise Accuracy (Pixel Acc.), and a metric $\Delta_{\text{AUC}}$ originally used by point tracking methods \citep{doersch2023tapir,karaev2023cotracker}, and later used for track point prediction by Track2Act. See Appendix \ref{apdx:metrics} for further details on metrics. 

Results are summarized in Table \ref{tab:track-prediction}, demonstrating that \modelname consistently leads to more accurate predictions, even though the forward dynamics model is only trained on a latent consistency loss rather than pixel-space prediction objectives. On the LIBERO dataset, we achieve over twice the pixel-wise accuracy of ATM, and we outperform Track2Act (which, unlike our method, has access to goal images) on their chosen $\Delta_{\text{AUC}}$ metric across BridgeData v2 and Something-Something v2. 
We attribute this success to several design choices, including the compression of motion into a compact latent space, thus improving efficiency and generalization; the prediction of discrete tokens to leverage the expressive power of autoregressive transformers; and the use of local-window pixel space classification, which gives our forward dynamics model the ability to model rich multi-modal distributions of motion and capture fine-grained dynamics. Further investigation into design choices (\ref{apdx:ablations}), detailed results (\ref{apdx:detailed-tables}), and qualitative visualizations (\ref{apdx:qualitative}) can be found in the Appendix.


\begin{table}[!t]
    \centering
    \small
    \caption{
        \textbf{Behavior Cloning} performance on LIBERO. \modelname is competitive with various state-of-the-art baselines, both with and without video pretraining.
    }
    \begin{tabular}{l c c c c c c}
        \toprule
        \textbf{Method} & \textbf{Video} & \textbf{LIBERO} & \textbf{LIBERO} & \textbf{LIBERO} & \textbf{LIBERO} & \textbf{LIBERO} \\
         & \textbf{Pre-training} & \textbf{Long} & \textbf{90} & \textbf{Object} & \textbf{Spatial} & \textbf{Goal} \\
        \midrule
        Diffusion Policy \citep{chi2023diffusion} & \xmark & 0.73 &  0.67 & \textbf{0.70} & 0.79 & 0.83 \\ 
        \rowcolor[HTML]{EFEFEF}
        QueST \citep{mete_quest_2024}             & \xmark & 0.67 & 0.89 & -- & -- & -- \\
        BAKU \citep{haldar2024baku}               & \xmark & \textbf{0.86} & \textbf{0.90} & -- & -- & -- \\
        \rowcolor[HTML]{EFEFEF}
        \modelname (Inverse only)                  & \xmark & 0.76 & 0.83 & 0.64 & \textbf{0.83} & \textbf{0.92} \\
        \midrule
        UniPi \citep{du_learning_2023}            & \checkmark & 0.06 & -- & 0.60 & 0.69 & 0.12 \\
        \rowcolor[HTML]{EFEFEF}
        ATM \citep{wen_any-point_2024}         & \checkmark & 0.44 &  0.63 & 0.81 & \textbf{0.79} & 0.59 \\
        \modelname (Full)                      & \checkmark & \textbf{0.75} & \textbf{0.88} &\textbf{ 0.93} & 0.73 & \textbf{0.92} \\
        \bottomrule
        \label{tab:libero-bc}
    \end{tabular}
    \vspace{-3mm}
\end{table}

\subsection{Utility of Predicted Latent Motions for Policy Learning}
\label{exp:policy}
\vspace{-2mm}
Beyond prediction accuracy, we examine whether video pre-training using \modelname can provide a useful prior for policy learning in both real-world and simulated experiments. Specifically, we evaluate \modelname along four dimensions measuring (1) in-distribution performance, (2) few-shot learning, (3) cross-embodiment transfer, and (4) generalization. Table \ref{tab:experiment-notation} summarizes the training datasets for different stages under each experimental setup. We evaluate performance using success rates on all five subsets of LIBERO, as well as a set of 3 real-world tasks: "Put the Rubik's Cube on the Box" (\texttt{Place Cube}), "Stack the Green and Blue Cups in the Orange Cup" (\texttt{Stack Cups}), and "Open the Box and Move the Eggplant into the Bowl" (\texttt{Open Box \& Place Eggplant})).

\paragraph{In-Distribution Performance}
We first evaluate \modelname in a standard behavior cloning setup, training both the forward and inverse dynamics models on only the demonstration data. We compare to state-of-the-art approaches with and without video pre-training. Results in Table \ref{tab:libero-bc} indicate that \modelname, even without additional data, is competitive with SOTA behavior cloning methods and outperforms video pre-training methods trained with (ATM) and without (UniPi) keypoint tracks. In this setting, we observe that since there is sufficient information to learn tasks to a high degree \textit{without} video pre-training, standard BC methods tend to match or outperform approaches using pre-training. However, in subsequent sections, we demonstrate that these approaches under-perform in limited data regimes and do not generalize effectively to new tasks.

\begin{figure}[!t] 
    \centering
    \includegraphics[width=\linewidth]{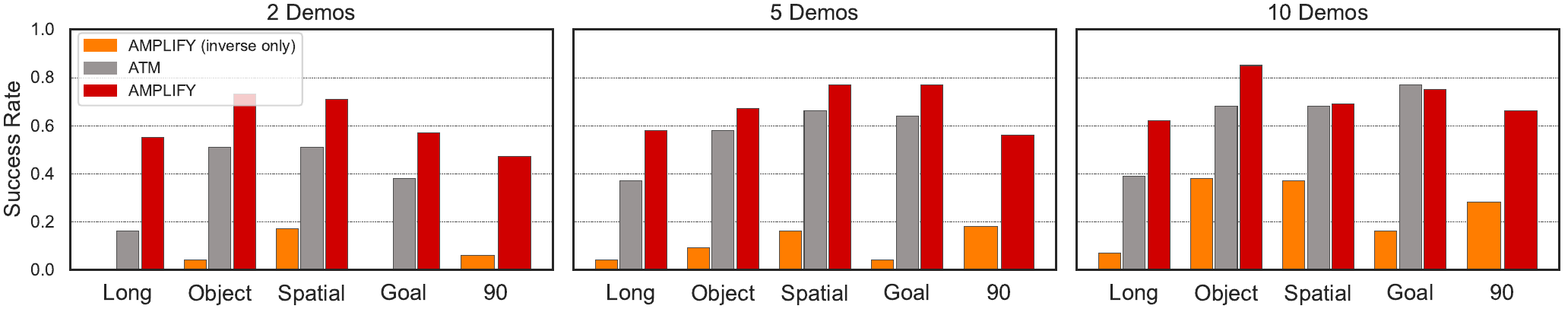}  
    \caption{\textbf{LIBERO few-shot}. Comparison of \modelname against ATM \citep{wen_any-point_2024} and a no-video-pre-training baseline. Our forward model is trained on all videos, and the inverse model is only trained on a limited number of demos. 
    } 
    \label{fig:few-shot-bar}
\end{figure}

\paragraph{Few-Shot Learning} 
We study whether \modelname can learn from fewer action-labeled demonstrations by training the forward model on all videos, while the inverse model is only trained on 4\%, 10\%, or 20\% of the 50 demonstrations available for each of the subsets of LIBERO.  In Figure \ref{fig:few-shot-bar}, we compare \modelname with ATM, trained on all videos and the same subsets of action data, as well as a variant of \modelname that does not condition on motion tokens to predict actions. Both \modelname and ATM consistently outperform the no-pre-training variant, indicating that in low-data regimes, video pre-training on keypoint dynamics provides a strong prior for data-efficient policy learning. In addition, \modelname achieves stronger performance than ATM on nearly every subset, suggesting that a latent motion representation has higher utility for action prediction than conditioning the policy directly on pixel-space track predictions. This seems to be especially true at the extreme low end--when provided with only 2 demonstrations per task, \modelname achieves an average $1.94\times$ improvement over ATM. Full numerical results are included in Table \ref{tab:libero-fewshot-full}.

\begin{table}[!b]
\centering
    \begin{minipage}{0.65\textwidth}
        \centering
        \resizebox{\linewidth}{!}{%
        \setlength{\tabcolsep}{3pt}
        \begin{tabular}{l c c c | c c c | c c c | c}
        \toprule
        Method & \multicolumn{3}{c}{Place Cube} & \multicolumn{3}{c}{Stack Cups} &  \multicolumn{3}{c}{Box/Eggplant}& Avg. \\
        \cmidrule(lr){2-4} \cmidrule(lr){5-7} \cmidrule(lr){8-10}
        \# Demos & 5 & 10 & All & 5 & 10 & All & 5 & 10 & All & \\
        \midrule
        Diffusion Policy \citep{chi2023diffusion}     & 0.6 & 0.5 & \textbf{0.9} & \textbf{0.3} & 0.5 & 0.5 & \textbf{0.1} & 0.2 & 0.2 & 0.42 \\
        \rowcolor[HTML]{EFEFEF}
        \modelname (DP head) & \textbf{0.7} & \textbf{0.9} & \textbf{0.9} & \textbf{0.3} & \textbf{0.6} & \textbf{1.0} & \textbf{0.1} & \textbf{0.3} & \textbf{0.4} & \textbf{0.58} \\
        \bottomrule
        \end{tabular}
        }
    \end{minipage}%
    ~
    \begin{minipage}{0.34\textwidth}
        \centering
        \caption{\textbf{Cross-Embodiment Transfer}. By leveraging human video demonstrations to train the forward dynamics model, \modelname outperforms Diffusion Policy on real-world tasks.}
        \label{tab:real-world-summary}
        \end{minipage}
        \vspace{-5pt}
\end{table}

\paragraph{Cross-Embodiment Transfer}
\label{sec:real-world}
Since the forward dynamics model can be trained on any observation data, we study whether videos of humans demonstrating a task can be used to improve policy learning. We train the forward dynamics model on both human and robot video data, while the inverse dynamics model is trained only on the action-labeled robot data. This setup highlights how the two stages can be decoupled to scale independently, unlike BC methods that cannot effectively harness action-free data. We evaluate success rates on three real-world tasks of varying difficulty, using Diffusion Policy as the BC baseline. For fair comparison, we replace the Gaussian head used in other experiments with a Diffusion Policy head in the inverse dynamics model. This ensures that the only difference between the two approaches is whether the predictions from our forward dynamics model are used to condition the policy. Similarly to the previous section, we evaluate \modelname in both the few-shot setting and the full demonstration setting. Results in Table \ref{tab:real-world-summary} demonstrate that \modelname can effectively leverage additional human data to learn common dynamics between human and robot motions, and use the predicted latent motions to improve policy learning. The average improvements of $1.32\times$, $1.4\times$, and $1.5\times$ indicate a more prominent gap as task complexity increases. See Table \ref{tab:real-world-full} for complete results.


\begin{table}[!t]
\centering    
    \begin{minipage}{0.6\textwidth}
        \centering
        \resizebox{0.99\linewidth}{!}{%
        \begin{tabular}{l c c c c}
            \toprule
            \textbf{Method} & \makecell{\textbf{LIBERO}\\\textbf{Long}} & \makecell{\textbf{LIBERO}\\\textbf{Object}} & \makecell{\textbf{LIBERO}\\\textbf{Spatial}} & \makecell{\textbf{LIBERO}\\\textbf{Goal}} \\
            \midrule
            Diffusion Policy \citep{chi2023diffusion}  & 0.00 & 0.00 & 0.00 & 0.00 \\
            \rowcolor[HTML]{EFEFEF}
            QueST \citep{mete_quest_2024}                & 0.07 & 0.00 & 0.01 & 0.01 \\
            BAKU \citep{haldar2024baku}                & 0.06 & 0.00 & 0.00 & 0.00 \\
            \rowcolor[HTML]{EFEFEF}
            \modelname (w/o tracks)      & 0.00 & 0.00 & 0.00 & 0.02 \\
            \modelname     & \textbf{0.52} & \textbf{0.80} & \textbf{0.69} & \textbf{0.41}  \\        
            \bottomrule
        \end{tabular}
        }        
    \end{minipage}%
    \hfill
    \begin{minipage}{0.39\textwidth}
        \centering
        \caption{\textbf{Zero-shot task generalization} from LIBERO 90 to unseen LIBERO subsets. We are the first to report non-trivial success on LIBERO without using \textit{any} action data from the target tasks. Compared to the best BC baseline, \modelname provides a $27\times$ average improvement. }  
        \label{tab:libero-generalization}
    \end{minipage}
\end{table}

\paragraph{Generalization}
\label{sec:generalization}
Observing that \modelname excels in learning from \textit{limited} action data, we now turn to a setting where \textit{no} action data is available for target tasks. Given \textit{only observations} of target tasks, as well as a dataset of out-of-distribution interaction data, we evaluate how well \modelname can solve the target tasks zero-shot. This challenging setting requires methods to both learn a good abstraction of the mapping from observations to actions, and also generalize that abstraction to predict correct actions on new tasks. To test this setting, we train the forward dynamics model on observations from all subsets of LIBERO, and train the inverse dynamics model and BC baselines on actions from \textit{only} LIBERO 90. We then evaluate on four LIBERO target suites (Long, Object, Spatial, Goal), specifically designed to test different categories of generalization \citep{liu2023libero}. We find that BC methods completely fail in this scenario, achieving near-zero success rates (Table \ref{tab:libero-generalization}). We attribute this failure to two main shortcomings of BC: (1) the supervised imitation objective has no incentive to learn a generalizable abstraction, and (2) BC has no mechanism for harnessing additional data that may be informative, such as videos. In contrast, \modelname attains an average 60.5\% success rate on target tasks, approaching the success rates of models that were directly trained on the target tasks. This success highlights the value of latent dynamics prediction as a versatile interface for learning general priors from action-free videos. In addition, it suggests that training a general reference following inverse dynamics model may be a more generalizable objective compared to imitation learning.


\subsection{Utility of Predicted Latent Motions for Conditional Video Generation}
\label{exp:video}
\vspace{-2mm}
To demonstrate the utility of predicting keypoint trajectories beyond robotic control, we condition a video prediction model \citep{Ko2023Learning} on the latent motion tokens predicted by our forward dynamics model. We find that conditioning a video prediction model on our latent motion tokens leads to improved generation quality (Table~\ref{tab:video-gen}). Compared to a baseline model that does not use track inputs, our approach yields better performance on all metrics (details on metrics in Appendix \ref{apdx:metrics}). This improvement suggests that our latent motion representation captures rich, structured dynamics that improve not only control tasks but also the fidelity of generated video content. Further details on training and generation are provided in Appendix \ref{apdx:training-video} and qualitative results in Appendix \ref{apdx:qualitative}.

\begin{table}[htb]
\centering
\small
\begin{tabular}{@{} c c @{}}
  \begin{minipage}{0.45\textwidth}
    \centering
    \resizebox{0.99\linewidth}{!}{%
      \begin{tabular}{l c c c}
        \toprule
        \textbf{Method} & \textbf{PSNR $\uparrow$} & \textbf{LPIPS $\uparrow$} & \textbf{SSIM $\uparrow$} \\
        \midrule
        AVDC \citep{Ko2023Learning} & 15.93 & 0.16 & 0.56 \\
        \rowcolor[HTML]{EFEFEF}
        AVDC + \modelname & \textbf{16.40} & \textbf{0.19} & \textbf{0.59} \\
        \bottomrule
      \end{tabular}
    }
  \end{minipage}
  \hfill
  \begin{minipage}{0.5\textwidth}
    \caption{Video Prediction. Conditioning AVDC on predicted motion tokens from our dynamics model improves generated video quality on BridgeData v2.}
    \label{tab:video-gen}    
  \end{minipage}
  \vspace{-3mm}
\end{tabular}
\end{table}


\section{Conclusion}
\label{conclusion}
\vspace{-2mm}
In this work, we introduced \modelname, a framework that leverages large-scale action-free video data and a small amount of interaction data to significantly enhance robotic policy performance. By decoupling the learning of \emph{what} constitutes a task from \emph{how} to execute it, our approach efficiently utilizes heterogeneous data sources. Our key insight lies in representing scene dynamics through compact latent motion tokens derived from keypoint trajectories, which enables higher efficiency and improved performance compared to pixel-level reconstruction methods. Experimental results show that \modelname consistently outperforms baseline methods, particularly in the limited action data regime and in zero-shot generalization settings. Moreover, the versatility of our latent representation extends beyond control, proving useful in tasks such as conditional video prediction. Our findings demonstrate the promise of harnessing large-scale human video data to inform robotic control policies and pave the way for more scalable, generalizable, and efficient robot learning.

\newpage 
\section{Limitations}
\label{limitations}
\vspace{-2mm}
While AMPLIFY is an exciting step towards robot learning from broad data sources, we recognize a number of limitations that could serve as promising directions for future research. First, by modeling tracks in 2D images, we are potentially leaving ambiguity in the inverse dynamics model if multiple actions could correspond to the same tracks. An explicitly 3D approach, predicting latent motions that correspond to 3D tracks \citep{SpatialTracker, ngo2024delta} could yield more robust representations that do not depend on fixed or known camera views. In addition, AMPLIFY currently only considers deterministic environment dynamics, since in stochastic settings additional information is required to separate agent actions from exogenous noise in purely state-to-state data \citep{misra2024towards, yang2022dichotomy, park2023hiql}. Since AMPLIFY has demonstrated the ability to learn from off-task data, it would also be interesting to explore whether the inverse dynamics model could be trained on data collected online by an exploration policy. Finally, future research could scale the prediction backbone to a general VLM or video prediction model to enhance video and language generalization.


\acknowledgments{If a paper is accepted, the final camera-ready version will (and probably should) include acknowledgments. All acknowledgments go at the end of the paper, including thanks to reviewers who gave useful comments, to colleagues who contributed to the ideas, and to funding agencies and corporate sponsors that provided financial support.}


\bibliography{main}  

\clearpage
\newpage
\appendix
\section*{Appendix}

In this document, we provide detailed supplementary material including a table summarizing notation (\ref{apdx:notation}), a discussion on the three-stage decomposition (\ref{apdx:3-stage}), extended related work (\ref{apdx:extended-related-work}), detailed experimental and training details (\ref{apdx:experiment-training-details}), ablation studies (\ref{apdx:ablations}), and additional quantitative (\ref{apdx:additional-results}) and qualitative (\ref{apdx:qualitative}) results.


\section{Notation}
\label{apdx:notation}
\vspace{-5mm}
\begin{table}[htbp]
\centering
\caption{Description of the Notation and Acronyms used in this manuscript}
\resizebox{\linewidth}{!}{%
\begin{tabular}{ll}
\toprule
\textbf{Symbol} & \textbf{Meaning} \\
\midrule
$ o_t $ & Image (visual) observation at time $t$. \\

$ q_t $ & Proprioceptive state of the robot at time $t$ (e.g., joint angles). \\
$ a_t $ & Action at time $t$. \\

$ g $ & Goal specification (e.g., language instruction or task label). \\
$ f $ & Forward dynamics model that autoregressively predicts latent motion tokens from $o_t$ and $g$. \\

$ f_{\mathrm{inv}} $ & Inverse dynamics model mapping latent motion tokens and current state $(o_t, q_t)$ to a sequence of actions. \\
$\pi$ & Our policy, defined as $f_{\mathrm{inv}}(o_t, q_t, f(o_t,g))$. \\

$\mathcal{V}$ & Video dataset $\{(o_t,g)\}$ \\
$\mathcal{R}$ & Action-labeled robot interaction dataset $\{(o_t,q_t,a_t)\}$\\

$ \kappa_t $ & Raw keypoint trajectories over a horizon from time $t$, with dimensions $T \times N \times 2$. \\
$ u_t $ & Single-timestep velocities computed from $\kappa_t$. \\

$ \tilde{z}_t $ & Continuous latent vectors produced by the keypoint encoder. \\
$ z_t $ & Discrete latent codes (tokens) representing keypoint motion, obtained via FSQ. \\

$ \mathcal{E}_\theta $ & Keypoint encoder that maps velocities $u_t$ to latent representations. \\
$ \mathcal{D}_\theta $ & Keypoint decoder that reconstructs velocity distributions from latent codes. \\

$ \omega_t $ & Ground-truth discretized labels for velocities, computed as $\Omega(u_t)$. \\
$ T $ & Prediction horizon (number of timesteps over which motion is predicted). \\

$ N $ & Number of keypoints in the grid. \\
$ W $ & Local window size for pixel classification in the decoder. \\

$ \bm{a}_t $ & Action chunk (sequence of actions over the horizon), i.e., $a_{t:t+T}$. \\
$ \gamma $ & Temporal discount factor used in the inverse dynamics loss. \\
\bottomrule
\end{tabular}
}
\label{tab:notation}
\end{table}


\section{Discussion on the Three-Stage Decomposition}
\label{apdx:3-stage}

One fundamental limitation of Behavior Cloning is that it is a monolithic architecture that requires paired (action, observation) data to learn a policy $\pi(o_t) = a_t$, which is not readily available at scale. Moreover, the data is assumed to be a goal-directed sequence of expert actions--standard imitation learning has no mechanism for harnessing interaction data that is suboptimal or not directed towards solving the tasks in the test set, even though such data (1) contains rich information about the relationship between visual observations, environment dynamics, and agent actions, and (2) may be easier to collect through exploration or play, compared to expert demonstrations \citep{wang2023mimicplay, chen_playfusion_2023, lynch_learning_2020}. In Section \ref{sec:generalization}, we demonstrate that these limitations prevent BC approaches from learning reusable abstractions. Based on these observations, we argue for a decoupled multi-stage approach that can learn from heterogeneous data sources. We classify data sources into three distinct categories based on their modality composition:

\textbf{Action-Free Videos}: observations of goal-directed behavior, but no action labels $\{(o_t,g)\}$

\textbf{Undirected Interaction Data}: observations and robot actions, but not in a goal-directed manner $\{(o_t,q_t,a_t)\}$

\textbf{Expert Robot Demonstrations}: goal-directed action-labeled rollouts $\{(o_t,q_t,a_t,g)\}$

Note that Expert Demonstrations can be treated as both Action-Free Videos and Interaction Data, since the modalities are a strict superset of the modalities in the other two. For this reason, in the main body of the paper we simply refer to $\mathcal{V}$ and $\mathcal{R}$, with any available demo data included in both sets by default. This taxonomy points towards one possible natural decomposition:
\begin{enumerate}
    \item Use Action-Free Videos (and Expert Demonstrations) to learn how observations evolve with respect to a goal
    \item Use Interaction Data (Undirected and Demonstrations) to learn how a sequence of observations maps to a sequence of actions
\end{enumerate}
This decomposition effectively decouples \textit{task understanding} (the sequence of observations that correspond to a goal) and \textit{task execution} (translating a reference sequence of states into low-level actions). Observations, the only shared modality, operate as the interface bridging the gap, serving as prediction targets for the first stage and input references for the second. Seeking a compact, action-informative representation of these observations, \modelname employs latent keypoint tokens, providing the third component of the three-stage decomposition. However, plenty of other representations are possible, including uncompressed images \citep{du_learning_2023} and pixel-space tracks \citep{wen_any-point_2024,bharadhwaj2024track2act}. In Algorithm \ref{algo:amplify} we summarize the training procedure for the three-stage approach, and in Table \ref{tab:data} we highlight the different data sources used for each component of \modelname in comparison to BC.

\begin{algorithm}[!t]
    \caption{\modelname Training.}
    \label{algo:amplify}
    \begin{algorithmic}[1]
        \Require Datasets $\mathcal{V}, \mathcal{R}$
        \State Preprocess keypoint tracks $\kappa_t$ in $\mathcal{V}$
        \State Learn latent motion encoding to compress $\kappa_t$ into discrete tokens $z_t$ using \Eq{autoencoder}
        \State Train forward dynamics model $f(o_t, g) = z_t$ on $\mathcal{V}$ using \Eq{dynamics}
        \State Train inverse dynamics model $f_{\text{inv}}(o_t, q_t, z_t) = a_{t:t+T}$ on $\mathcal{R}$ using \Eq{inv_dynamics}
        \State \Return $\pi(o_t, q_t, g) = f_{\text{inv}}(o_t, q_t, f(o_t, g))$
    \end{algorithmic}
\end{algorithm}

\begin{table}[h]
    \centering
    \caption{Compared to BC, AMPLIFY can leverage video and off-task data by decoupling forward and inverse dynamics.}
    \label{tab:data}
    \resizebox{0.7\linewidth}{!}{%
    \begin{tabular}{l c c c c}
        \toprule
        \textbf{Data type} & \text{BC} & \makecell{Forward\\Dynamics} & \makecell{Inverse\\Dynamics} & \text{\modelname} \\
        \midrule
        Action-Free Videos                &        &  \cmark &        & \cmark \\
        
        Expert Robot Demonstrations             & \cmark &  \cmark & \cmark & \cmark \\
        Undirected Interaction Data  &        &         & \cmark & \cmark \\
        \bottomrule
    \end{tabular}
    }
    \vspace{3mm}
\end{table}


\section{Extended Related Work}
\label{apdx:extended-related-work}

In this section we provide context on additional related works and alternative approaches for robot learning from videos.


\subsection{Learning from Hand Pose}
Videos have shown to be an effective source of data for learning robotic policies from human demonstrations. One method for attaining action labels for unannotated videos is to estimate hand pose to gain information about human action. \citep{wang2023mimicplay, bharadhwaj2023zero, kareer2024egomimic} estimate the trajectory of the hand position or pose, and then train a policy to replicate these trajectories with a robotic arm. While this representation reduces the domain gap, it lacks granularity, as the degrees of freedom represented (either 3 or 6 DoF) fall short of capturing the full complexity of the human hand, which possesses 20-30 DoF.

The retargeting of human actions to the robot's action space is another prevalent strategy. This has been achieved through various means, including image translation architectures \citep{xiong2021learning} analytical remappings to optimize cost functions \citep{shaw2023videodex, mandikal2022dexvip}, and masking the agent from the scene \citep{bahl2022human}.

\subsection{Learning from Affordances}
Affordances, or the set of ways in which a given object or environment may be manipulated, are a common abstraction between human and robot data that lends itself well to learning from videos. The estimation of contact locations, trajectories, and future states are prevalent strategies for interpreting and acting upon environmental cues \citep{mandikal2022dexvip, bahl2023affordances, mendonca2023structured}. These methods aim to deduce actionable information from video data, and attempt to learn how to interact with various objects and environments based on observed human actions.

\subsection{Reward/Representation Learning from Videos}
A common method of extracting action-relevant information from videos is via self-supervised representation learning.
Some works align video data with language descriptors via contrastive learning \citep{nair2022r3m, nair2022learning} or minimize the distance from a specified goal representation \citep{ma2022vip}. Some works extract latent action representations from videos such as \citep{schmeckpeper2020reinforcement, bhateja2023robotic, ghosh2023reinforcement}. More recently, works such as \citep{bruce_genie_2024, chen_moto_2024, ye2024latent} extract latent actions via pixel-level reconstruction and use them to learn from action-free videos.

Representations learned from video and language often serve as the basis for reward or value functions in deep reinforcement learning settings, predominantly within goal-conditioned RL frameworks \citep{ma2022vip, schmeckpeper2020reinforcement, ghosh2023reinforcement}, where the aim is to produce actions that minimize the distance to the desired outcome. Other works such as \citep{escontrela2023video, huang2023diffusion} utilize models trained on objectives such as video prediction to estimate values.

\subsection{Forward and Inverse Dynamics for Robot Learning}

\modelname benefits from multiple sources of data by decoupling the problem of policy learning into forward and inverse dynamics, where the forward dynamics model predicts future states or latent representations, and the inverse dynamics model maps these predictions to actions. \citep{cui2024dynamo, lapo} focus on recovering latent actions from video data using self-supervised pretraining, enabling control with minimal action labels. Methods such as \citep{wen2024vidmanexploitingimplicitdynamics, liang2024dreamitaterealworldvisuomotorpolicy, Ko2023Learning, du_learning_2023} use text-guided video generation to predict future visual trajectories, from which actions can be inferred. These works present a promising direction for transferring information from large-scale video data into visuomotor policies using pre-trained foundation and frontier models, thus improving generalization to new tasks and environments.



\section{Experimental and Training Details}
\label{apdx:experiment-training-details}

In this section we provide extensive details on the LIBERO benchmark (\ref{apdx:benchmark-details}), our real-world setup (\ref{apdx:real-details}), metrics used for evaluation (\ref{apdx:metrics}), preprocessing (\ref{apdx:preprocessing}), and training details for each stage (\ref{apdx:training}). 
     \begin{figure}[p]  
        \centering
        \includegraphics[width=0.9\columnwidth]{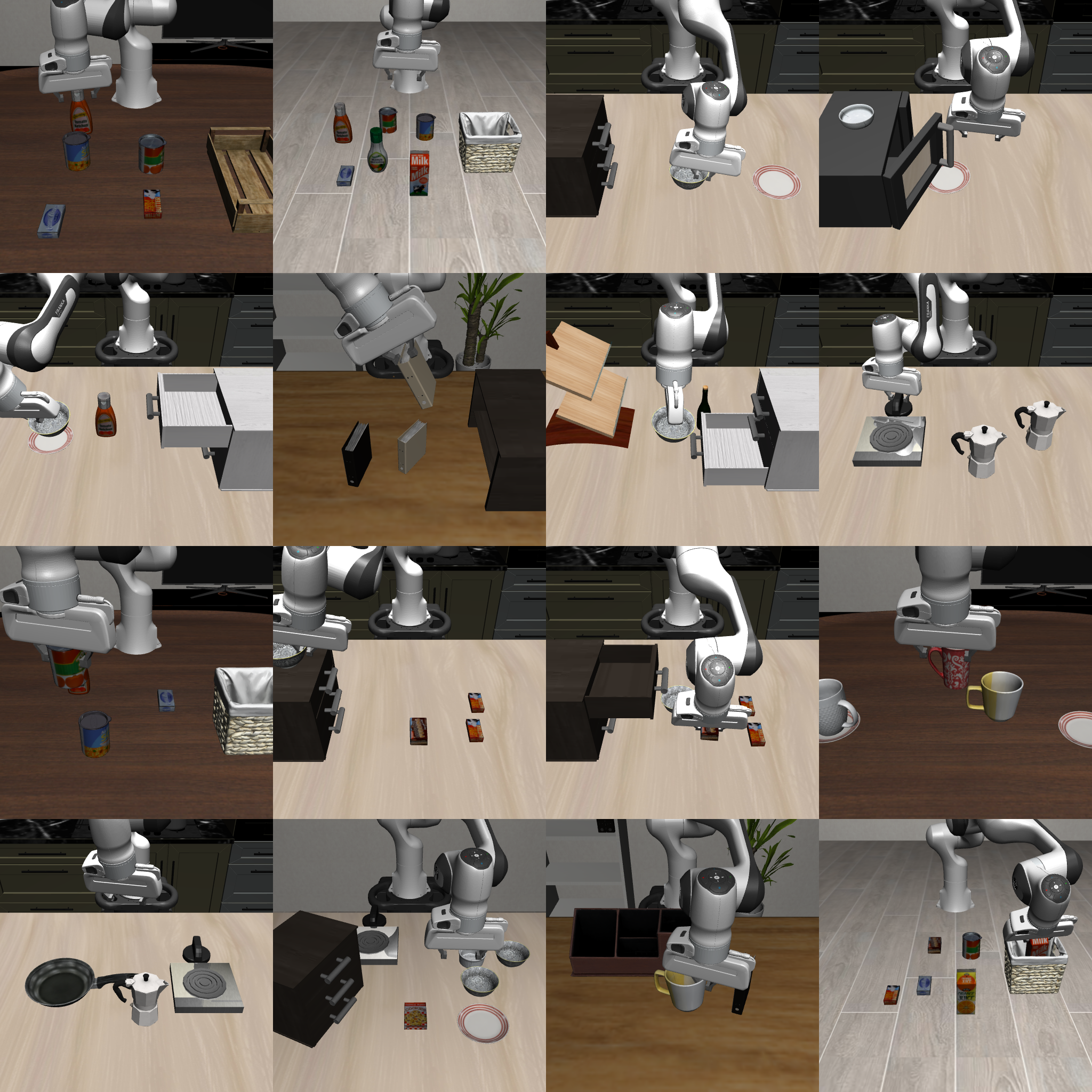} 
        \caption{A sample of the 130 diverse tasks and environment configurations in LIBERO.}
        \label{fig:libero-tasks-image}
    \end{figure}
    
    \subsection{LIBERO Benchmark}
    \label{apdx:benchmark-details}

    We evaluate on the LIBERO \citep{liu2023libero} benchmark, which consists of 130 manipulation tasks. The LIBERO benchmark is categorized into distinct subsets:
    \begin{itemize}
        \item \textbf{LIBERO-Long}: A subset of 10 long-horizon manipulation tasks.
        \item \textbf{LIBERO-90}: A broad set of 90 tasks with diverse objects, layouts, and backgrounds.
        \item \textbf{LIBERO-Object}: Tasks that evaluate generalization to novel object categories.
        \item \textbf{LIBERO-Spatial}: Tasks that test generalization across varied spatial arrangements.
        \item \textbf{LIBERO-Goal}: Tasks with the same starting scene but different goals to assess goal conditioning.
    \end{itemize}
    Figure \ref{fig:libero-tasks-image} shows a sample of the tasks and environments in the collection. The benchmark comes with a dataset of 50 expert demonstrations per task, obtained through VR teleoperation \citep{liu2023libero}. When evaluating on LIBERO, \modelname takes in the standard $128\times128$ RGB images as obserations and produces normalized axis-angle actions. We execute actions in the environment at 20 Hz, and give our model a maximum of 500 environment steps to solve each task. We perform rollouts on 10 random seeds per task for all subsets of LIBERO except for LIBERO-90, for which we only perform one rollout for each of the 90 tasks to produce our results.


    \subsection{Real-World Setup}
    \label{apdx:real-details}

    \begin{figure}[p]  
        \centering
        \includegraphics[width=\columnwidth]{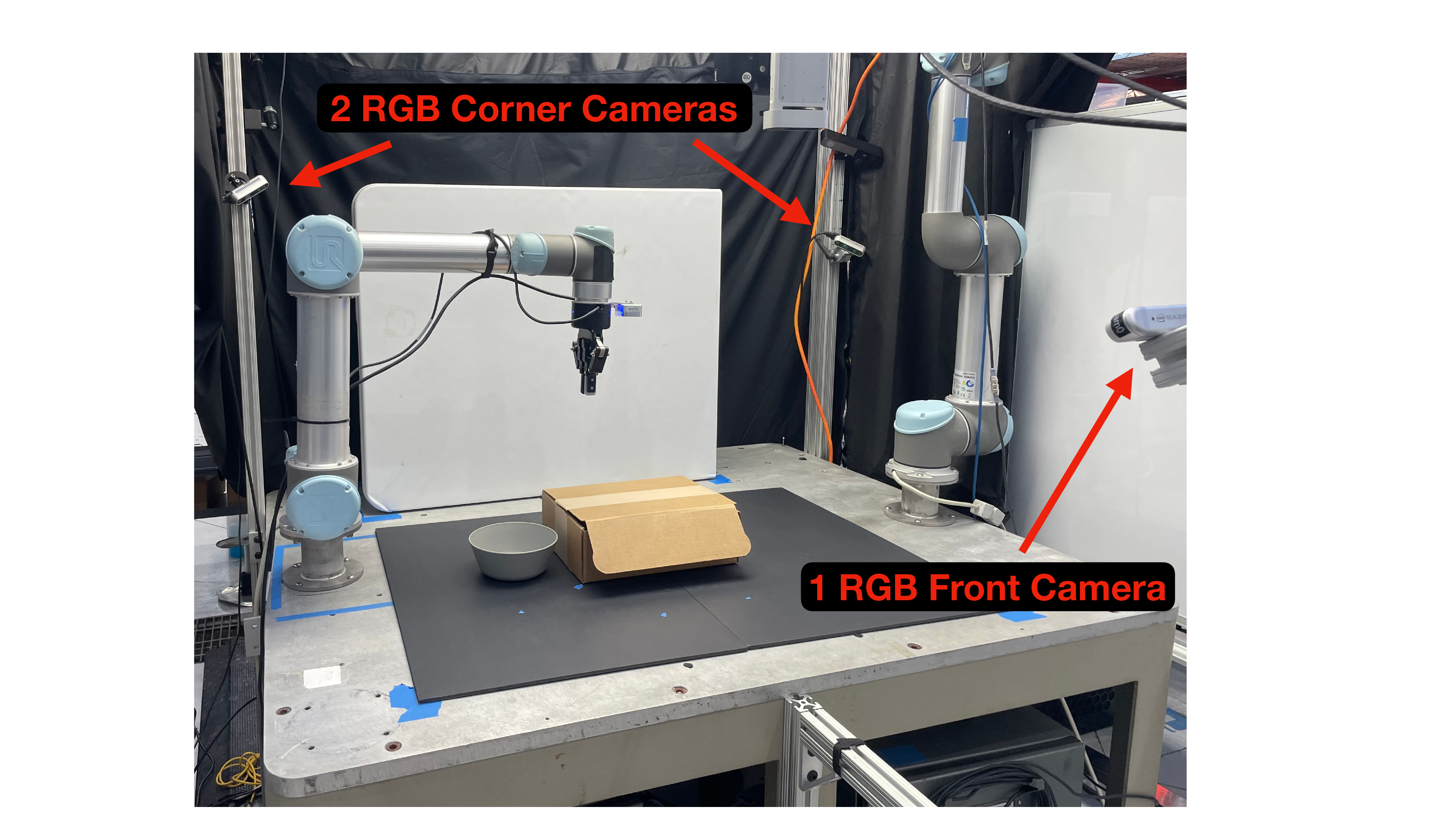}  
        \caption{We use three static RGB cameras as input observations for both human and robot (UR5) data.}
        \label{fig:real-setup}
    \end{figure}
    
    \begin{figure}[p]
        \centering
        
        \includegraphics[width=\linewidth]{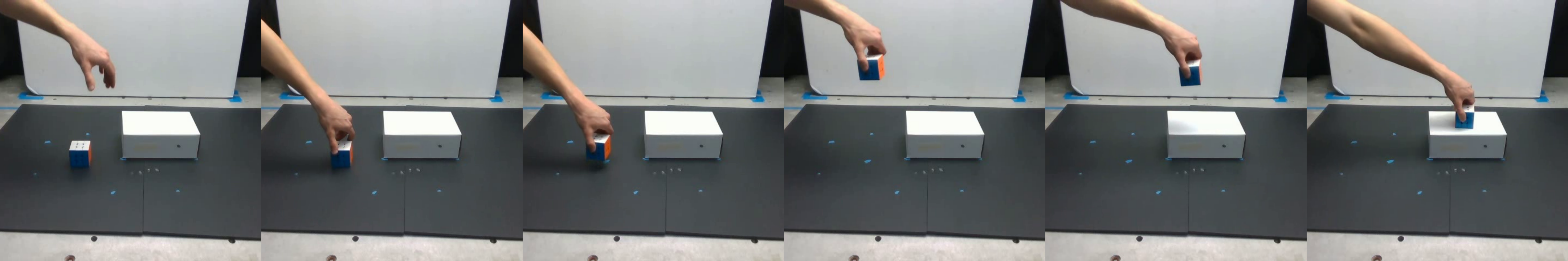}
        \includegraphics[width=\linewidth]{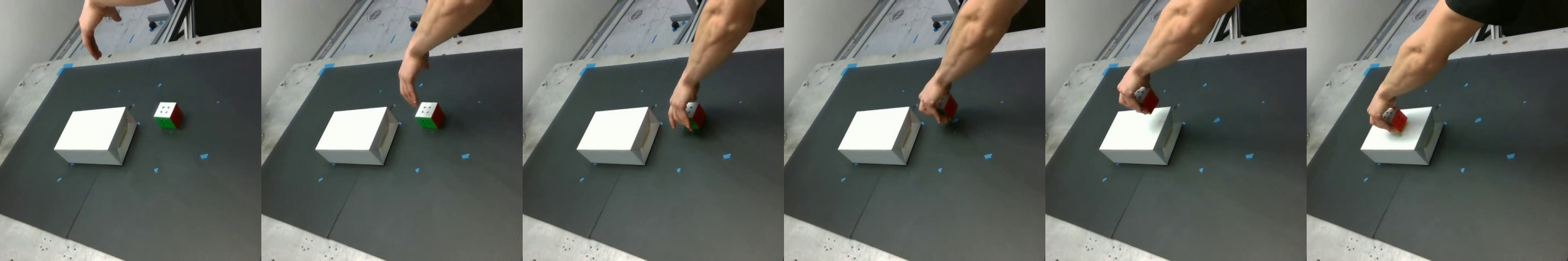}
        \includegraphics[width=\linewidth]{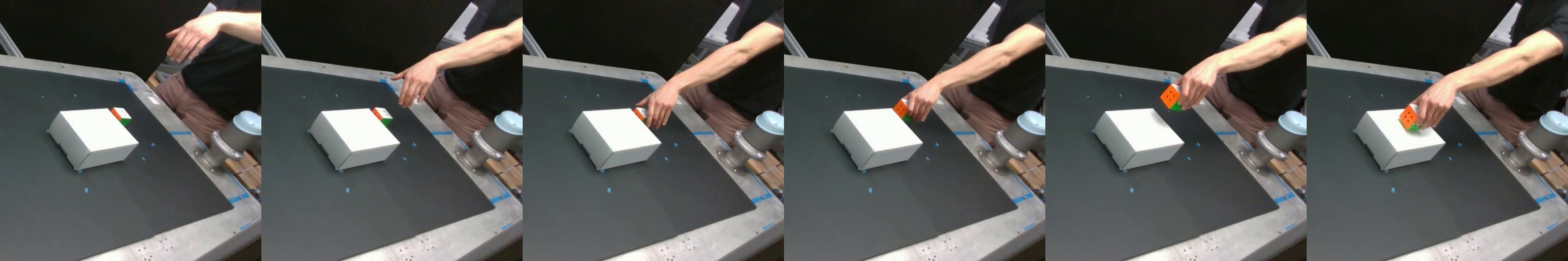}
        \vspace{1mm}
        
        \includegraphics[width=\linewidth]{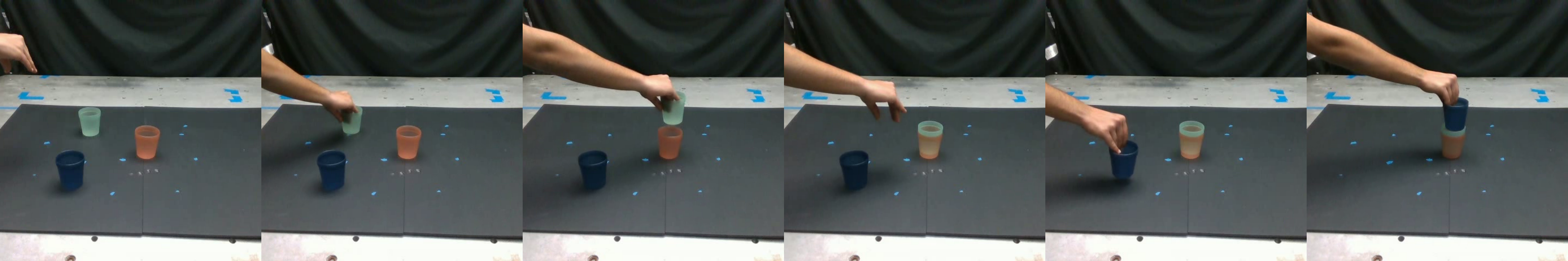}
        \includegraphics[width=\linewidth]{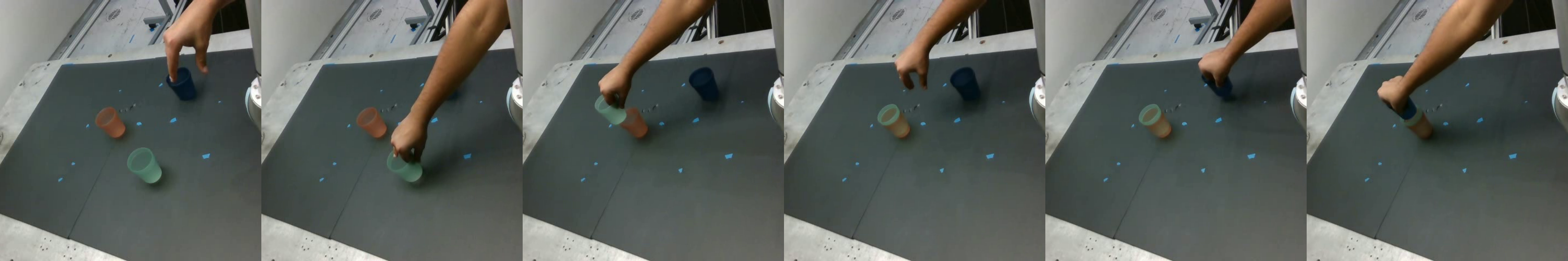}
        \includegraphics[width=\linewidth]{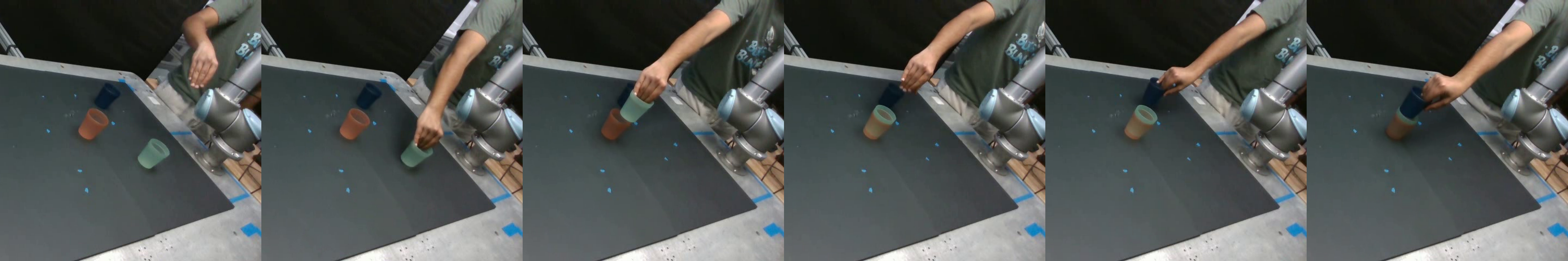}
        \vspace{1mm}
        
        \includegraphics[width=\linewidth]{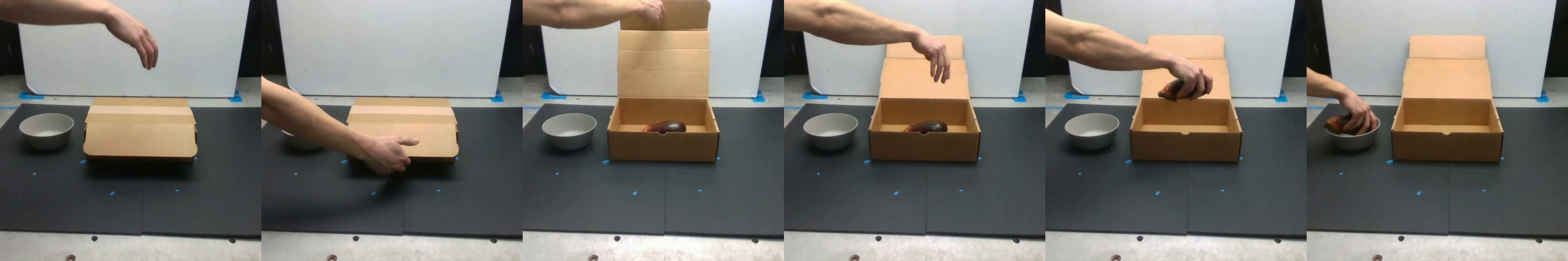}
        \includegraphics[width=\linewidth]{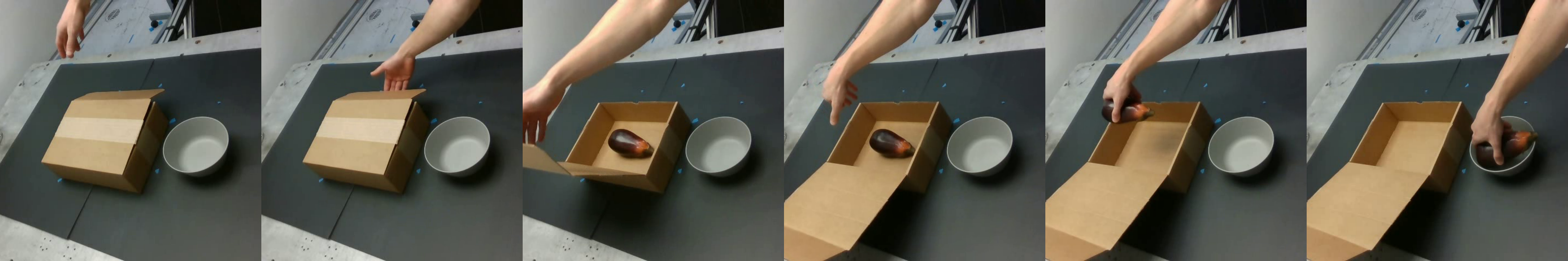}
        \includegraphics[width=\linewidth]{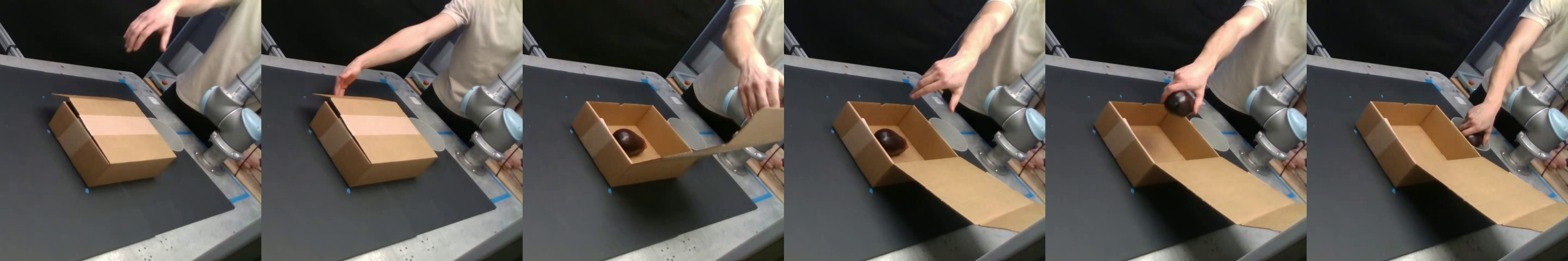}
        
        \caption{\textbf{Real-World Tasks and Cameraviews.}
        Each row shows front and corner camera views for three different tasks: 
        \textit{Task 1:} “Put the Rubik's Cube on the Box”, 
        \textit{Task 2:} “Stack the Green and Blue Cups in the Orange Cup”, and 
        \textit{Task 3:} “Open the Box and Move the Eggplant into the Bowl”.}
        \label{fig:real-tasks-cameras}
    \end{figure}

    \paragraph{Robot and Camera Setup}
    In our real-world experiments, we evaluate the performance of our method on three distinct manipulation tasks using a UR5 robotic arm equipped with three synchronized static RGB camera views positioned on three sides of the workspace (Pictured in Figure \ref{fig:real-setup}). This multi-view configuration ensures robust perception of the scene. We do not use a wrist camera. Camera observations are captured at 60Hz but downsampled to 20Hz to match action inputs on the robot. Images are cropped and resized to $224\times224$ for all views. Actions are output as absolute normalized end-effector positions, which we found to work better than delta pose.

    \paragraph{Task Descriptions and Demonstration Details}
    To test performance on our real-world setup, we evaluate \modelname across three tasks of varying difficulty. Fig \ref{fig:real-tasks-cameras} shows each task from the 3 cameraviews, and Table \ref{tab:demo-counts} records the number of human/robot demonstrations we collect for each task.
    \begin{itemize}
        \item \textbf{Task 1: "Put the Rubik's Cube on the Box"} - This is the easiest task, consisting of a single pick-and-place movement. We slightly vary the initial placement and orientation of the cube during both demonstrations and evaluation. 
        \item \textbf{Task 2: "Stack the Green and Blue Cups in the Orange Cup"} - This tasks consists of two cup-stacking motions, requiring more precise grasping and a longer-horizon rollout. We measure both partial and full success rate (stacking one or both cups correctly, invariant of order) and report full results in Table \ref{tab:real-world-full}. Cup placements are also varied slightly during demonstrations and evaluations.
        \item \textbf{Task 3: "Open the Box and Move the Eggplant into the Bowl"} - This task requires precise object articulation (the clearance between the box lid and the table is only a few centimeters), continued contact, and then a pick-and-place movement with partial observability. Like Task 2, we measure both partial and full success rates (partial for opening the box, full for also placing the eggplant in the bowl correctly), also shown in Table \ref{tab:real-world-full}. We vary the orientation of the eggplant and position of the box flap.
    \end{itemize}

    \begin{table}[h]
        \centering
        \small
        \caption{\textbf{Demonstration Counts per Task.} Number of robot and human demonstrations collected for each real-world task.}
        \label{tab:demo-counts}
        \begin{tabular}{lcc}
            \toprule
            \textbf{Task} & \textbf{Robot Demos} & \textbf{Human Demos} \\
            \midrule
            \texttt{Place Cube} & 24 & 48 \\
            
            \texttt{Stack Cups} & 13 & 59 \\
            \texttt{Open Box \& Place Eggplant} & 15 & 30 \\
            \bottomrule
        \end{tabular}
    \end{table}
    
    

    \paragraph{Evaluation Criteria}
    We measure success rates from 10 rollouts on each task for both \modelname and the baseline Diffusion Policy, with a time limit of 90 seconds. Partial and full success rates are measured as follows:

    \textbf{\texttt{Place Cube}}
    \vspace{-5pt}
        \begin{itemize}
            \item \textit{Partial Success:} N/A.
            \item \textit{Full Success:} The Rubik’s cube is safely placed on the box.
        \end{itemize}

    \textbf{\texttt{Stack Cups}}
    \vspace{-5pt}
        \begin{itemize}
            \item \textit{Partial Success:} Two out of the three cups are successfully stacked.
            \item \textit{Full Success:} All three cups are correctly stacked.
        \end{itemize}

    \textbf{\texttt{Open Box \& Place Eggplant}}
    \vspace{-5pt}
        \begin{itemize}
            \item \textit{Partial Success:} The box is opened, regardless of whether the eggplant is picked and placed.
            \item \textit{Full Success:} The robot opens the box and successfully places the eggplant in the bowl.
        \end{itemize}

    
    \subsection{Metrics}
    \label{apdx:metrics}

    \paragraph{Track Prediction Metrics}
    We use three main metrics to measure performance in our keypoint trajectory (track) prediction experiments: mean squared error (MSE), pixel accuracy, and $\Delta_{\text{AUC}}$ (Area Under Curve). For each, we use tracks obtained from CoTracker \citep{karaev2023cotracker} as "ground-truth" for predictions. All methods use a uniformly spaced $20\times20$ grid as initial query points to track. Track2Act \citep{bharadhwaj2024track2act} also used this initial grid for their predictions, so we take their reported numbers directly. In ATM \citep{wen_any-point_2024}, the model is trained to take in any point location (including the uniform grid). Seer \citep{gu2023seer} is a video prediction model, so any queries can be applied to its output and tracked by CoTracker.
    \begin{enumerate}
        \item \textbf{MSE}: We take normalized track predictions in the range $[-1,1]$ and compute MSE between predicted $(x,y)$ values for each point and the corresponding ground-truth point from CoTracker.
        \item \textbf{Pixel Accuracy}: We measure the (normalized) percentage of predictions that are pixel-perfect compared to ground-truth.
        \item \textbf{$\Delta_{\text{AUC}}$}: This metric was originally introduced by works presenting point tracking \citep{doersch2023tapir,karaev2023cotracker} and later used for track prediction \citep{bharadhwaj2024track2act}. The metric is computed as follows. Let $\delta_t^x$ be the fraction of point predictions that are within a threshold pixel distance of $x$ of their ground truth in a time-step $t \in [0,H]$. Following \citep{bharadhwaj2024track2act}, we report the area under the curve $\Delta$ with $\delta_t^x$ by varying $x$ from $1$ to $N=10$ and taking the average across the prediction horizon $H$ i.e. $\Delta=\left(\sum_{t=1}^H \sum_{x=1}^N \delta_t^x\right) / H$. Hence, $\Delta$ is normalized to $[0,1]$, with higher values corresponding to better predictions.
    \end{enumerate}

    \paragraph{Video Prediction Metrics}
    We use three standard metrics for measuring generated video quality: Learned Perceptual Image Patch Similarity (LPIPS), Structural Similarity Index (SSIM), and Peak Signal-to-Noise Ratio (PSNR). These metrics are defined as follows:
    \begin{align*}
        \text{LPIPS}(x, \hat{x}) &\coloneq \frac{1}{HW} \sum_{h,w} ||\phi(x)_{hw} - \phi(\hat{x})_{hw}||_2^2 \\
        \text{SSIM}(x, \hat{x}) &\coloneq \frac{(2\mu_x\mu{\hat{x}} + c_1)(2\sigma_{x\hat{x}} + c_2)}{(\mu_x^2 + \mu_{\hat{x}}^2 + c_1)(\sigma_x^2 + \sigma_{\hat{x}}^2 + c_2)} \\
        \text{PSNR}(x, \hat{x}) &\coloneq 10 \log_{10}\left(\frac{\mathrm{MAX}_I^2}{\frac{1}{HW}\sum_{h,w}(x_{hw}-\hat{x}_{hw})^2}\right) \\
    \end{align*}
    \noindent where $x_{hw}$ and $\hat{x}_{hw}$ are ground truth and predicted images respectively, $\phi$ denotes a pretrained deep network,  $\mu$ and $\sigma$ represent mean and variance, $c_1$ and $c_2$ are constants for numerical stability, and $\mathrm{MAX}$ is the maximum possible pixel value. Note that Frechet Inception Distance (FID) and Frechet Video Distance (FVD) are not meaningful metrics in our scenario due to a mismatch between the input image size for the Inception network and the videos produced by AVDC. 

    
    \subsection{Preprocessing}
    \label{apdx:preprocessing}
    We unfold (in time) a length-$\tau$ video $o\in\mathbb{R}^{\tau\times H\times W\times 3}$ into $\tau$ length-$T$ windows to be tracked by CoTracker \citep{karaev2023cotracker}. For each window, we initialize a uniform $N=20\times20$ grid of query points to be tracked through the $T$ frames of the window. Thus, we obtain an output $\kappa_{0:\tau} \in \mathbb{R}^{\tau\times T\times N\times2}$, where $\kappa_t$ are the tracks corresponding to the grid initialized in $o_t$ and tracked through $o_{t:t+T}$. 
    
    Note that tracks corresponding to nearby windows overlap in time, but correspond to \textit{different} initial query points, since we re-initialize in every frame. This is in contrast to the way tracks are preprocessed in ATM \citep{wen_any-point_2024}, where the points are initialized \textit{once} (in the last frame) and the \textit{same} points are tracked through the entire video. Our preprocessing strategy is $T\times$ more compute-intensive, but prevents issues of prolonged occlusion and gracefully handles moving/panning cameras, which is important when using a wrist camera, for example.

    Once tracks are obtained from CoTracker, we treat them as ground-truth targets for the rest of the pipeline. For training stability, we normalize tracks from pixel coordinates to the range $[-1,1]$. To ensure tracks correspond to the same length of time, we interpolate tracks in time on same datasets that have a different sampling frequency. On LIBERO, Real robot tasks, and Something-Someting v2 we use a 16-frame true horizon, corresponding to 0.8 seconds (at 20Hz). For BridgeData v2, the observation frequency is 5Hz so we interpolate a 4-frame horizon to 16 using a cubic spline. For real-world human data, we use 8 frames, since human execution is roughly $2\times$ that of the robot. In Appendix \ref{apdx:ablations} we examine the effect of changing prediction horizon on downstream policy performance and find that 16 frames (0.8s) lead to the best result. This echoes findings in \citep{wen_any-point_2024}. Before tokenizing, we compute instantaneous velocities through simple differencing in the time dimension. Since the initial queries are always the same, velocities contain the same information and reduce the sequence length slightly.
    
    
    \subsection{Training Details}
    \label{apdx:training}

    We present detailed hyperparameter choices in Table \ref{tab:hyperparameters} and a high-level overview of AMPLIFY training in Algorithm \ref{algo:amplify}. All three components were trained on a single GPU (either RTX 6000 or L40S, depending on availability). We report the default number of epochs used for experiments. For the LIBERO benchmark, we trained the motion tokenizer for approximately 50,000 gradient steps, the forward dynamics model for 25,000 gradient steps, and the inverse dynamics model for 75,000 gradient steps. For Something-Something v2, we trained the motion tokenizer for 32,000 gradient steps and the forward dynamics model for 22,000 gradient steps. For BridgeData, we trained the motion tokenizer for 20,000 gradient steps and the forward dynamics model for 175,000 gradient steps.
    Each gradient step is counted as the accumulated gradients over 4 backward passes to simulate the reported batch size.
    
    \begin{table}[h]
        \centering
        \caption{Hyperparameters for Motion Tokenizer, Forward Dynamics, and Inverse Dynamics model training.}
        \label{tab:hyperparameters}
        \begin{tabular}{lccc}
            \toprule
            Hyperparameter & Motion Tokenizer & Forward Dynamics & Inverse Dynamics \\
            \midrule
            number of parameters & 31M & 70M & 57M \\
            
            epochs & 100 & 100 & 250 \\
            batch size & 256 & 256 & 256 \\
            
            gradient accumulation & 4 & 4 & 4 \\
            learning rate & 1e-4 & 1e-4 & 1e-4 \\
            
            optimizer & AdamW & AdamW & AdamW \\
            image size & - & $128\times 128$ & $128\times 128$ \\
            
            number of points $N$ & - & 400 & - \\
            track prediction horizon $T$ & - & 16 & - \\
            
            decoder local window size $W$ & - & 15 & - \\
            number of heads & 8 & 8 & 8 \\
            
            number of layers & 2 & 8 & 4 \\
            hidden dimension & 768 & 768 & 768 \\
            
            dropout & 0.1 & 0.1 & 0.1 \\
            FSQ implicit codebook size & - & 2048 & - \\
            
            Action loss discount $\gamma$ & - & - & 0.99 \\
            \bottomrule
        \end{tabular}
    \end{table}
        
        \paragraph{Video Prediction}
        \label{apdx:training-video}
        To demonstrate the utility of our method beyond robotics, we extend the AVDC \citep{Ko2023Learning} video generation model by conditioning it on the motion tokens produced by the forward dynamics model. These motion tokens serve as a conditioning signal that guides video prediction based on the expected dynamics. To condition the video generation model, motion tokens generated by the forward dynamics model are concatenated with text tokens along the sequence dimension before features before being pooled by a Perceiver \citep{jaegle_perceiver_2021} resampler in AVDC's UNet. To ensure efficient batching during training and inference, the concatenated sequences are padded to a fixed sequence length. During training, the video prediction model is conditioned on motion tokens obtained directly from the motion tokenizer. This direct conditioning allows the model to learn directly from ``ground truth" motion tokens, increasing training efficiency.
        
        At test time, a new sequence of motion tokens is sampled every $T$ timesteps. This sampling strategy allows the model to generate a complete trajectory by sequentially updating the conditioning information over time.
        The model is conditioned on the outputs of the forward dynamics model during inference, similarly to \modelname's inverse dynamics model. We find that track-conditioned video generation leverages the strength of our forward dynamics model to guide the video generation process, thus improving the quality and coherence of the generated video sequences.

\section{Ablation Studies}
\label{apdx:ablations}

We conduct an ablation analysis to study the effect of various design choices in our architecture. All evaluations are performed on LIBERO-Long, which is the most challenging subset of LIBERO due to it's long horizon (up to 500+ steps) compared to the other tasks.

    \paragraph{Motion Tokenization}
    \label{apdx:ablations-tokenization}
    We examine the effect of a number of architectural choices in the Motion Tokenizaton stage, using the $\Delta_{\text{AUC}}$ as the principal metric, summarized in Table \ref{tab:ablations-tokenization}. 
    \begin{itemize}
        \item \textbf{Attention Masks}: We consider three attention masks in the encoder: per-timestep tokenization, where there is no information transfer across time; causal attention, where information flows one way in time, and no mask. We observe a significant benefit from attention across time, but marginal difference between causal and full attention (we opt to use causal for efficiency).
        \item \textbf{Decoder Output Loss}: We consider two options for decoder output loss: MSE, and Local Window Classification. Under the MSE setup, the decoder is configured to directly output (normalized) pixel coordinates, which are regressed to the targets from CoTracker. With Local Window Classification, the decoder instead outputs classification logits over $W^2=225$ classes for each point for each timestep. Each class corresponds to a pixel in a the local $W^2$ window of pixels centered around the previous point in the current timestep. For example, the class corresponding to the middle of the window predicts a velocity (0,0), whereas the class corresponding to the top-right pixel is (7,7). The size of the window can be inferred from the data and leads to a bias for local motion. We observe that Local Window Classification models tracks better and leads to more accurate predictions. Qualitatively, we notice that MSE tends to lead to blurrier predictions and tends towards 0-movement more. We suspect this is because the classification objective can model multi-modal distributions, whereas MSE simply regresses to the mean.
        \item \textbf{Joint Tokenization}: We study whether conditioning track tokenization on the image helps reconstruction by conditioning the encoder on both image tokens (same as forward dynamics) and the input velocities. Performance dropped slightly, though not much difference was apparent.
        \item \textbf{Prediction Horizon}: We vary the prediction horizon from 4 to 16 timesteps and find that reconstructing tracks over a shorter horizon is easier. This is expected, as there is less uncertainty with shorter horizons. We observe a similar case when trainig forward dynamics. However, the inverse dynamics model has the opposite trend, since longer horizon predictions are ultimately more useful for action inference. As a result, we choose to use a horizon of 16 in our final model, as we are ultimately interested in downstream policy performance.
        \item \textbf{Model Dimensions}: Finally, we examine FSQ (effective) codebook size, code sequence length, and hidden dimension of the transformers. We find that an effective codebook size of 2048, a latent sequence length of 16, and a hidden dimension of 768 perform marginally better.
    \end{itemize}

    \begin{table*}[!p]
        \centering
        \small
        \caption{Motion Tokenizer Ablations.}
        \label{tab:ablations-tokenization}
        \begin{tabular}{l | l c c}
            \toprule
            \textbf{Ablation Factor} & \textbf{Configuration} & \textbf{Metric} & \textbf{Performance} \\
            \midrule
            \multirow{3}{*}{Attention Mask} 
            & Per-Timestep & $\Delta_{\text{AUC}}$ & 0.877 \\
            Attention Mask & \textbf{Causal} & $\Delta_{\text{AUC}}$ & \textbf{0.919} \\
            & Full & $\Delta_{\text{AUC}}$ & 0.918 \\
    
            \midrule
            \multirow{2}{*}{Decoder Output Loss} 
            & MSE Loss & $\Delta_{\text{AUC}}$ & 0.883 \\
            &\textbf{Local Window Classification Loss} & $\Delta_{\text{AUC}}$ & \textbf{0.919} \\
    
            \midrule
            \multirow{2}{*}{Joint Tokenization} 
            & \textbf{Tracks Only} & $\Delta_{\text{AUC}}$ & \textbf{0.929} \\
            & Tracks + Image & $\Delta_{\text{AUC}}$ & 0.926 \\
    
            \midrule
            \multirow{3}{*}{Prediction Horizon} 
            & \textbf{4 }& $\Delta_{\text{AUC}}$ & \textbf{0.985} \\
            & 8 & $\Delta_{\text{AUC}}$ & 0.961 \\
            & 16 & $\Delta_{\text{AUC}}$ & 0.919 \\
    
            \midrule
            \multirow{3}{*}{Codebook Size} 
            & 512 & $\Delta_{\text{AUC}}$ & 0.912 \\
            Codebook Size & 1024 & $\Delta_{\text{AUC}}$ & 0.919 \\
            & \textbf{2048 }& $\Delta_{\text{AUC}}$ &\textbf{ 0.921} \\

            \midrule
            \multirow{5}{*}{Hidden Dimension} 
            & 128 & $\Delta_{\text{AUC}}$ & 0.897 \\
            & 256 & $\Delta_{\text{AUC}}$ & 0.909 \\
            Hidden Dimension
            & 384 & $\Delta_{\text{AUC}}$ & 0.911 \\
            & 512 & $\Delta_{\text{AUC}}$ & 0.914 \\
            & \textbf{768} & $\Delta_{\text{AUC}}$ & \textbf{0.919} \\
            & 1024 & $\Delta_{\text{AUC}}$ & 0.917 \\
    
            \midrule
            \multirow{5}{*}{Code Sequence Length} 
            & 2 & $\Delta_{\text{AUC}}$ & 0.853 \\
            & 4 & $\Delta_{\text{AUC}}$ & 0.877 \\
            Code Sequence Length & 8 & $\Delta_{\text{AUC}}$ & 0.909 \\
            & \textbf{16} & $\Delta_{\text{AUC}}$ & \textbf{0.919} \\
            & 32 & $\Delta_{\text{AUC}}$ & 0.893 \\

            \bottomrule
        \end{tabular}
        \vspace{-5mm}
    \end{table*}

    \paragraph{Forward Dynamics}
    \label{apdx:ablations-forward} 
    We study design choices in the forward dynamics model, which predicts tokenized trajectories from observations. We evaluate models based on $\Delta_{\text{AUC}}$ and pixel accuracy, summarized in Table \ref{tab:ablations-forward}.
    
    \begin{itemize} 
        \item \textbf{Prediction Horizon}: Similar to motion tokenization, we observe that shorter prediction horizons improve accuracy. Predicting only 4 steps ahead achieves the highest accuracy, while 16-step prediction is substantially harder. We nevertheless use 16 in the final model to match the inverse dynamics setup. 
        \item \textbf{Vision Encoder Architecture}: We evaluate multiple vision encoders. Interestingly, despite the popularity of larger and pretrained architectures (e.g., DINOv2 \citep{oquab2023dinov2}, \citep{dosovitskiy2020image}), ResNet-18 \citep{veit_residual_2016} performs competitively, with minimal drop in performance, making it a computationally efficient default choice. 
        \item \textbf{Token Pooling Strategy}: We compare using CLS tokens vs. patch tokens from ResNet-18 as inputs to the transformer. Patch tokens (i.e., per-patch embeddings) slightly underperform CLS pooling, but are preferable due to their richer spatial structure and compatibility with other modules. 
        \item \textbf{Transformer Depth}: We evaluate transformer depth and find that using 4 layers performs slightly better than 8. This may be due to overfitting or optimization instability with deeper models on limited data. 
        \item \textbf{Frozen vs. Fine-tuned Vision Encoder}: We find no benefit to fine-tuning the ResNet encoder during forward dynamics training, so we freeze it to save compute and stabilize training. 
    \end{itemize}

    \begin{table*}[!p]
        \centering
        \small
        \caption{Forward Dynamics Ablations.}
        \label{tab:ablations-forward}
        \begin{tabular}{l | l c c}
            \toprule
            \textbf{Ablation Factor} & \textbf{Configuration} & \textbf{Metric} & \textbf{Performance} \\
            \midrule
            \multirow{3}{*}{Prediction Horizon}
            & \textbf{4} & Pixel Accuracy & \textbf{0.757} \\
            
            Prediction Horizon & 8 & Pixel Accuracy & 0.678 \\
            & 16 & Pixel Accuracy & 0.613 \\
            
            \midrule
            \multirow{3}{*}{Vision Encoder}
            & \textbf{ResNet-18} & Pixel Accuracy & \textbf{0.613} \\
            
            Vision Encoder & ResNet-50 & Pixel Accuracy & 0.621 \\
            & DINOv2 & Pixel Accuracy & 0.621 \\
            
            & ViT & Pixel Accuracy & 0.614 \\
            
            \midrule
            \multirow{1}{*}{Token Pooling Strategy}
            & \textbf{Patch Tokens} & Pixel Accuracy & \textbf{0.613} \\
            
            & CLS Token & Pixel Accuracy & 0.621 \\
            
            \midrule
            \multirow{1}{*}{Transformer Depth}
            & \textbf{4 Layers} & $\Delta_{\text{AUC}}$ & \textbf{0.930} \\
            
            & 8 Layers & $\Delta_{\text{AUC}}$ & 0.929 \\
            
            \midrule
            \multirow{1}{*}{Vision Encoder Tuning}
            & \textbf{Frozen} & $\Delta_{\text{AUC}}$ & \textbf{0.929} \\
            
            & Fine-tuned & $\Delta_{\text{AUC}}$ & 0.929 \\
            
            \bottomrule
        \end{tabular}
        \vspace{-5mm}
    \end{table*}

    \begin{table}[!p]
        \centering
        \small
        \caption{Inverse Dynamics Ablations.}
        \label{tab:ablations-inverse}
        \begin{tabular}{l | l c c}
            \toprule
            \textbf{Ablation Factor} & \textbf{Configuration} & \textbf{Metric} & \textbf{Performance} \\
            \midrule
            & 4 & Success Rate & 0.36 \\
            
            Prediction Horizon & 8 & Success Rate & 0.64 \\
            & \textbf{16} & Success Rate & \textbf{0.75} \\
            \midrule
            & \textbf{Gaussian (Transformer, MLP)} & Success Rate & \textbf{0.74} \\
            
            Action Head & \textbf{Diffusion (U-Net)} & Success Rate & \textbf{0.74 }\\
            & Flow Matching (DiT) & Success Rate & 0.73 \\
            \bottomrule
        \end{tabular}
        \vspace{-5mm}
    \end{table}

    \paragraph{Inverse Dynamics}
    \label{apdx:ablations-inverse} 
    Finally, we investigate the impact of the action prediction horizon and choice of output head in the inverse dynamics module, using downstream task success rate as the evaluation metric. Results are shown in Table \ref{tab:ablations-inverse}.
    \begin{itemize} 
        \item \textbf{Prediction Horizon}: We observe a consistent improvement in task success as we increase the action prediction horizon. This makes intuitive sense: longer horizons provide more context for disambiguating latent trajectories and allow the inverse dynamics model to recover the intended actions more reliably. We thus adopt a 16-step horizon in our final setup. 
        \item \textbf{Action Head}: We experiment with three different action heads. (1) a Gaussian Action head, consisting of a transformer decoder and MLP projection to output mean and log-std for a Gaussian Policy, trained with negatice likelihood loss as described in the main paper; (2) a Diffusion Policy head with U-net architecture (recommended for simple tasks), which we use in the real world setup for fair comparison. (3) A Flow Matching \citep{lipman2023flowmatchinggenerativemodeling} head with the cross-attention DiT architecture from GR00T N1 \citep{nvidia2025gr00tn1openfoundation}, with optimal transport coupling and 10 integration steps with a midpoint ODE solver for inference. Our evaluations did not highlight significant differences in performance, so we opted for the simplest Gaussian Policy for the main results. This follows \citep{haldar2024baku}, who similarly found that a complex action head did not help on LIBERO tasks.
    \end{itemize}

\section{Additional Results}
\label{apdx:additional-results}


    \subsection{Video Scaling Experiment}
    \label{apdx:video-scaling}
    \begin{table}[!htbp]
        \centering
        \caption{We scale the amount of video data used \modelname, showing that performance improves as we scale the amount of action-free video data. Note that these policies are trained on only 2 action-annotated trajectories.}
        \label{tab:video-scaling}
        \resizebox{0.5\linewidth}{!}{
            \begin{tabular}{l c c c c c}
                \toprule
                \textbf{\# Videos} & \makecell{\textbf{0}} & \makecell{\textbf{2}} & \makecell{\textbf{5}} & \makecell{\textbf{10}} & \makecell{\textbf{50}} \\
                \midrule
                Ours   & 0.00 & 0.12 & 0.34 & 0.23 & 0.55 \\
                \bottomrule
            \end{tabular}
        }
        \vspace{-5pt}
    \end{table}
    
    In order to investigate the impact of abundant action-free video data on policy performance, we conducted an experiment on the LIBERO-Object subset. In this setup, we vary the number of videos used to train the forward dynamics model while keeping the action-annotated dataset extremely limited (only 2 trajectories). The goal is to simulate a realistic scenario where acquiring action labels is expensive, yet large-scale video data is readily available.
    
    Specifically, we trained the forward dynamics model using 0, 2, 5, 10, and 50 action-free video clips of LIBERO rollouts. Subsequently, the inverse dynamics model was trained using the outputs of the forward dynamics model as described in Section \ref{sec:prediction}. The results summarized in Table \ref{tab:video-scaling} demonstrate that policy performance generally improves as the volume of video data increases. It is worth noting that while the overall trend is positive, there is a non-monotonic behavior (e.g., a drop from 0.34 with 5 videos to 0.23 with 10 videos). This variation could be due to inherent stochasticity in training or differences in video content quality. In addition, since all of the demonstrations are likely similar, the forward dynamics model may not gain significantly more information as the number of demonstrations increase. Overall, these findings suggest that augmenting limited action-annotated data with large-scale, action-free video data can effectively improve the learned forward dynamics, which in turn improves policy performance. However, more thorough investigation is needed before drawing definitive conclusions.

    
    \subsection{Detailed Tables of Main Results}
    \label{apdx:detailed-tables}
    We provide more detailed versions of tables provided in the main body. 
    %

    \begin{table*}[!htbp]
        \centering
        \small
        \caption{Prediction Performance on Track Prediction Metrics on all LIBERO subsets.}
        \label{tab:libero-prediction-full}
        \resizebox{\linewidth}{!}{
            \begin{tabular}{llcccccc}
                \toprule
                \textbf{Method} & \textbf{Metric} & \shortstack{\textbf{LIBERO}\\\textbf{90}} & \shortstack{\textbf{LIBERO}\\\textbf{Long}} & \shortstack{\textbf{LIBERO}\\\textbf{Object}} & \shortstack{\textbf{LIBERO}\\\textbf{Spatial}} & \shortstack{\textbf{LIBERO}\\\textbf{Goal}} & \textbf{Aggregate} \\
                \midrule
                \multirow{3}{*}{ATM \citep{wen_any-point_2024}} 
                & MSE                       & 0.012 & 0.008 & 0.008 & 0.074 & 0.009 & 0.022 \\
                
                ATM \citep{wen_any-point_2024} & $\Delta_{\text{AUC}}$     & 0.799 & 0.803 & 0.782 & 0.693 & 0.757 & 0.767 \\
                & Pixel Accuracy            & 0.339 & 0.349 & 0.222 & 0.146 & 0.195 & 0.250 \\
                \midrule
                \multirow{3}{*}{\modelname} 
                
                & MSE                       & 0.004 & 0.002 & 0.001 & 0.019 & 0.002 & 0.006 \\
                & $\Delta_{\text{AUC}}$     & 0.892 & 0.921 & 0.937 & 0.904 & 0.914 & 0.913 \\
                
                & Pixel Accuracy            & 0.612 & 0.633 & 0.656 & 0.613 & 0.630 & 0.629 \\
                \bottomrule
            \end{tabular}
        }
        \vspace{-5mm}
    \end{table*}

    \begin{table}[!htbp]
        \centering
        \small
        \caption{Few-shot Learning from Limited Action Data on LIBERO. }
        \label{tab:libero-fewshot-full}
        \resizebox{\linewidth}{!}{
            \begin{tabular}{l*{16}{c}}
                \toprule
                Method           & \multicolumn{3}{c}{LIBERO-Long} & \multicolumn{3}{c}{LIBERO-90} & \multicolumn{3}{c}{LIBERO-Object} & \multicolumn{3}{c}{LIBERO-Spatial} & \multicolumn{3}{c}{LIBERO-Goal} \\
                \cmidrule(lr){2-4} \cmidrule(lr){5-7} \cmidrule(lr){8-10} \cmidrule(lr){11-13} \cmidrule(lr){14-16}
                \# Demos         & 2 & 5 & 10 & 2 & 5 & 10 & 2 & 5 & 10 & 2 & 5 & 10 & 2 & 5 & 10 \\
                \midrule
                ATM \citep{wen_any-point_2024}              & 0.16  & 0.37  & 0.39  & --    & --    & --    & 0.51  & 0.58  & 0.68  & 0.51  & 0.66  & 0.68  & 0.38  & 0.64  & \textbf{0.77} \\
                
                \modelname (inverse only)  & 0.00  & 0.04  & 0.07  & 0.06    & 0.18    & 0.28    & 0.04  & 0.09  & 0.38  & 0.17  & 0.16  & 0.37  & 0.00     & 0.04  & 0.16 \\
                \addlinespace[0.5ex]
                \modelname    & \textbf{0.55} & \textbf{0.58} & \textbf{0.62} & \textbf{0.47} & \textbf{0.56} & \textbf{0.66} & \textbf{0.73} & \textbf{0.67} & \textbf{0.85} & \textbf{0.71} & \textbf{0.77} & \textbf{0.69} & \textbf{0.57} & \textbf{0.77} & 0.75\\
                \bottomrule
            \end{tabular}
        }
    \end{table}

    \begin{table*}[!htbp]
        \centering
        \small
        \caption{Real-World Task Performance with Partial and Full Success Rates.}
        \label{tab:real-world-full}
        \resizebox{\linewidth}{!}{
            \begin{tabular}{ccccccccccccccccc}
                \toprule
                Method & \multicolumn{3}{c}{\makecell{Open Box \& Place\\ Eggplant - Partial}} & \multicolumn{3}{c}{\makecell{Open Box \& Place \\  Eggplant - Full}} & \multicolumn{3}{c}{\makecell{Stack Cups \\ Partial}} & \multicolumn{3}{c}{\makecell{Stack Cups \\ Full}} & \multicolumn{3}{c}{Place Cube} & Avg \\
                \cmidrule(lr){2-4} \cmidrule(lr){5-7} \cmidrule(lr){8-10} \cmidrule(lr){11-13} \cmidrule(lr){14-16}
                \# Demos & 5 & 10 & All & 5 & 10 & All & 5 & 10 & All & 5 & 10 & All & 5 & 10 & All & \\
                \midrule
                DP \citep{chi_diffusion_2023}     & \textbf{0.5} & 0.2 & 0.3 &\textbf{ 0.1} & 0.2 & 0.2 & \textbf{0.8} & 0.8 & 0.9 & \textbf{0.3} & 0.5 & 0.5 & 0.6 & 0.5 & \textbf{0.9} & 0.49 \\
                
                \modelname & 0.1 & \textbf{0.4} & \textbf{0.5 }& \textbf{0.1} & \textbf{0.3} & \textbf{0.4} & \textbf{0.8} & \textbf{0.9} &\textbf{ 1.0 }&\textbf{ 0.3} & \textbf{0.6} & \textbf{1.0} & \textbf{0.7} & \textbf{0.9} &\textbf{ 0.9} & \textbf{0.59} \\
                \bottomrule
            \end{tabular}
        }
    \end{table*}

    
    \subsection{Qualitative Results}
    \label{apdx:qualitative}
    We provide qualitative results of predictions from \modelname and samples from the video prediction model conditioned on predicted tracks. Within each frame, yellow indicates points at the current time, and red indicates the points in the future. Model outputs are for the full 400-point grid. To reduce clutter, however, we do not visualize points that are predicted to have no motion.

    \begin{figure}[p]  
        \centering
        \includegraphics[width=\columnwidth]{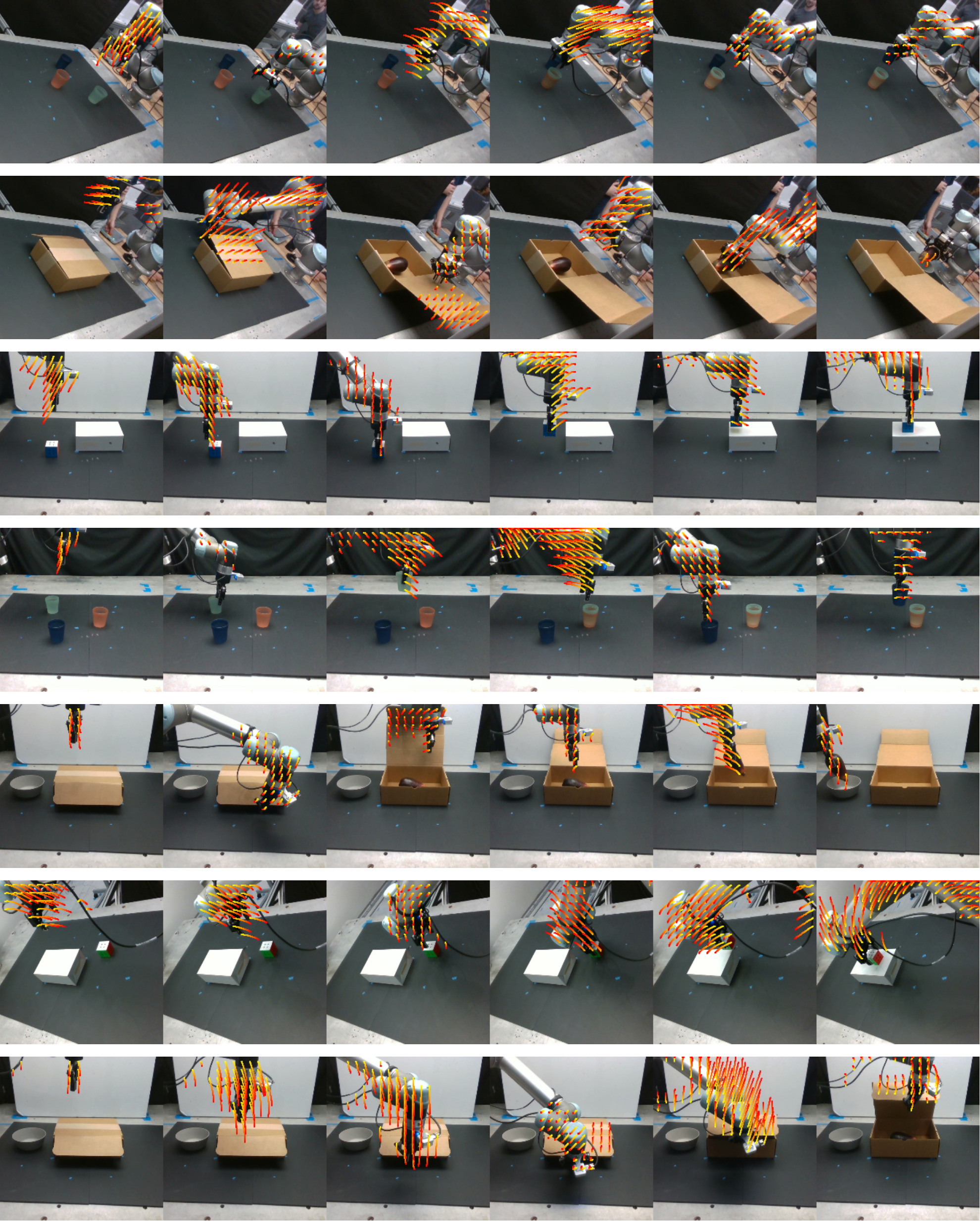}  
        \caption{Track Predictions from \modelname on Real-World Robot Data. }
        \label{fig:robot-qualitative}
    \end{figure}

    \begin{figure}[p]  
        \centering
        \includegraphics[width=\columnwidth]{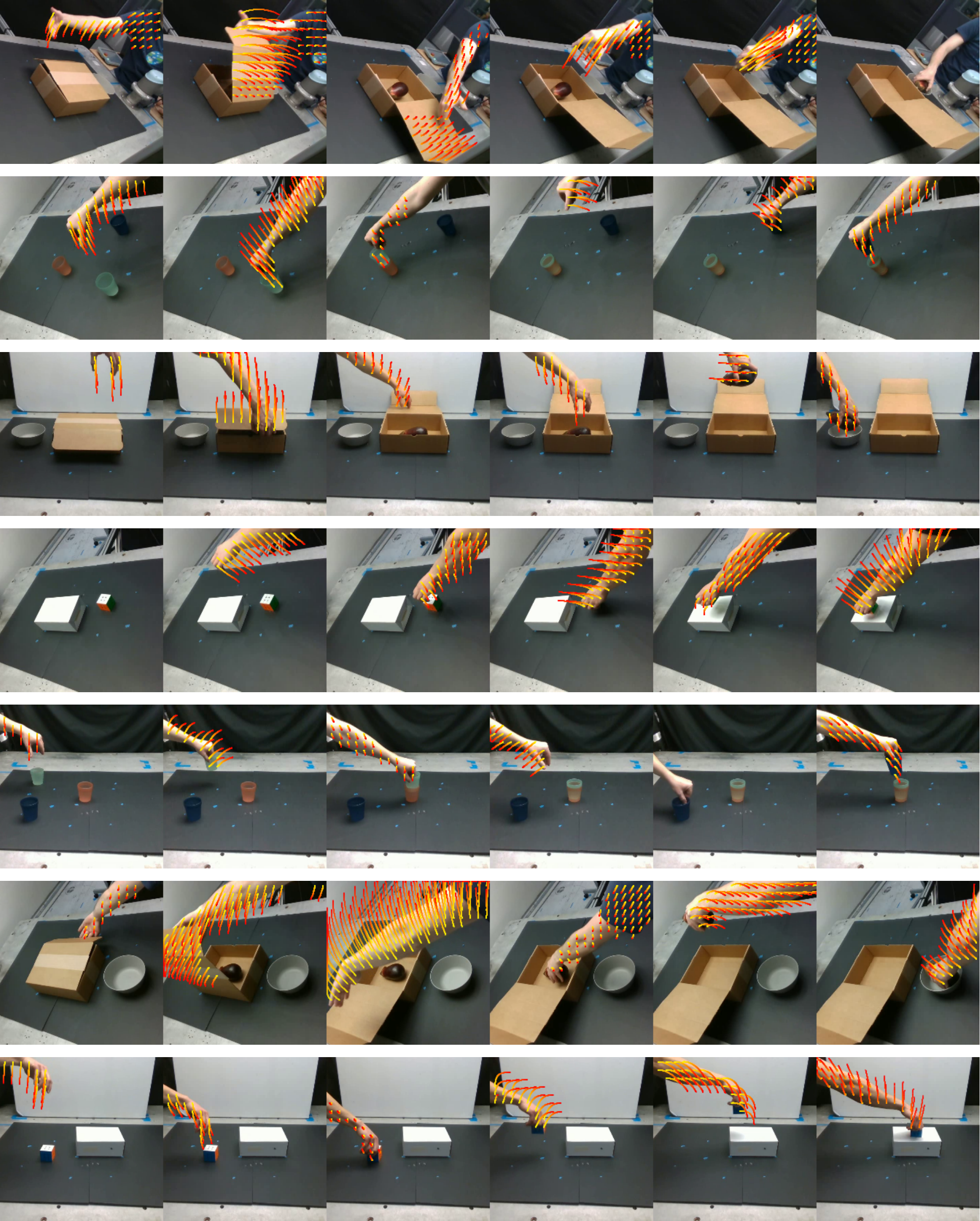}  
        \caption{Track Predictions from \modelname on Real-World Human Data. }
        \label{fig:human-qualitative}
    \end{figure}

    \begin{figure}[p]  
        \centering
        \includegraphics[width=\columnwidth]{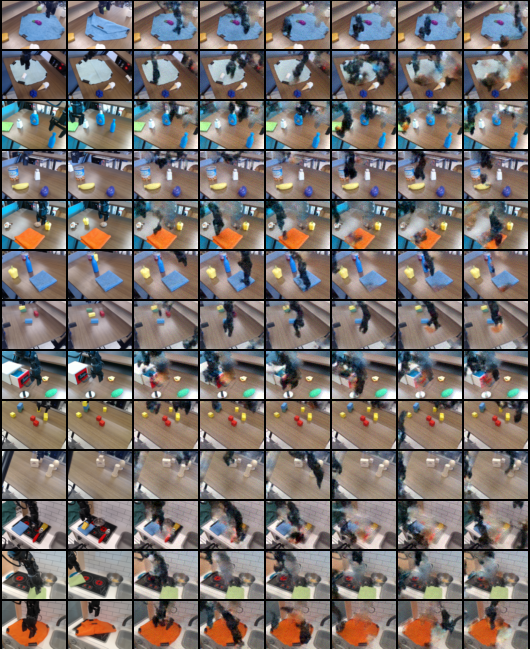}  
        \caption{Video Predictions from AVDC \citep{Ko2023Learning} conditioned on \modelname, trained on Bridge Data. }
        \label{fig:video-pred-sample}
    \end{figure}

    \begin{figure}[p]  
        \centering
        \includegraphics[width=\columnwidth]{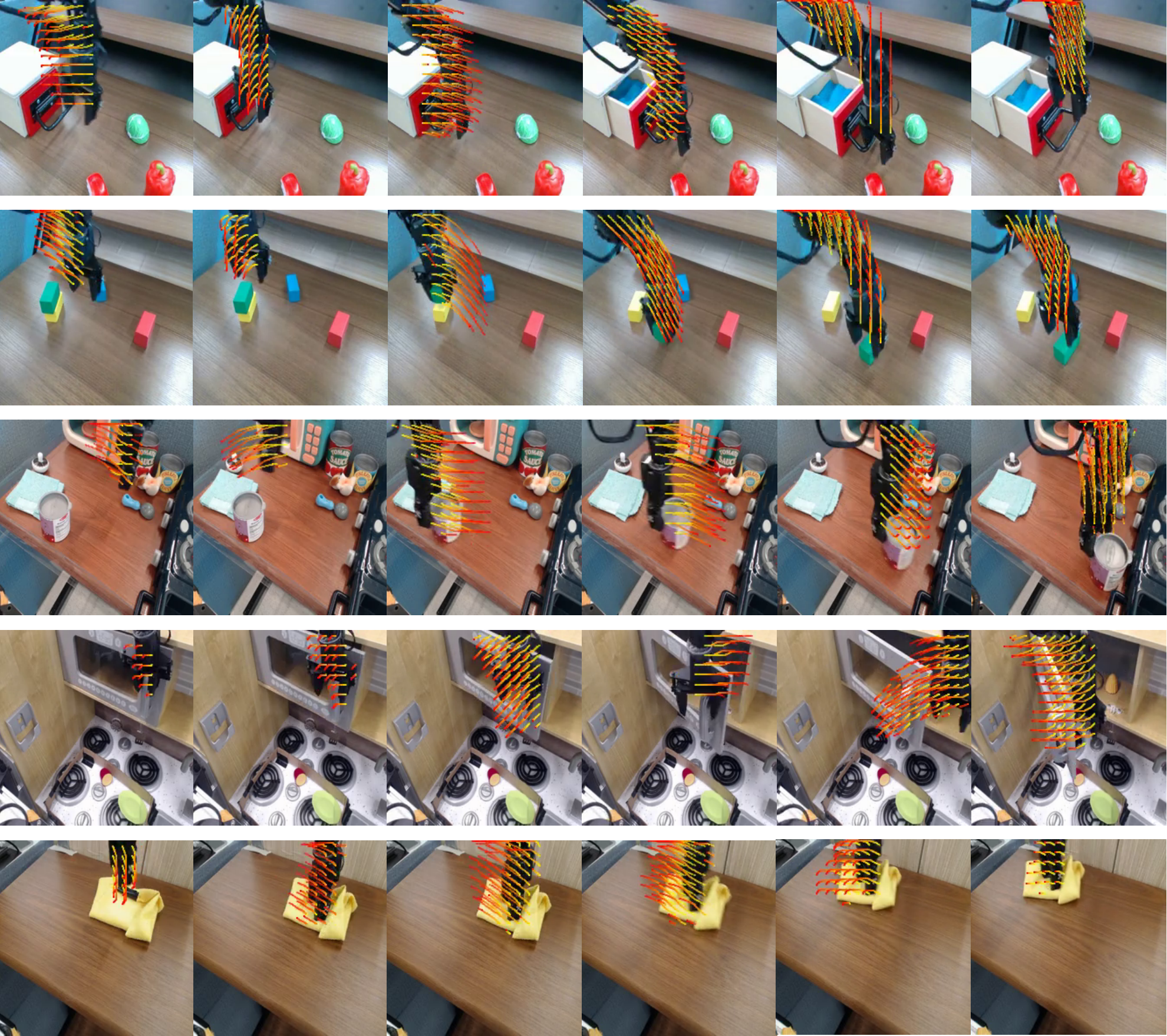}  
        \caption{Track Predictions from \modelname on Bridge Data. }
        \label{fig:bridge-qualitative}
    \end{figure}





\end{document}